\documentclass[acmtog,screen,nonacm]{acmart}

\usepackage{hyperref}

\usepackage{graphicx}
\usepackage{subcaption}
\usepackage[capitalize]{cleveref}

\usepackage{orcidlink}

\usepackage{wrapfig}
\usepackage{multirow}
\usepackage{tabularx}
\usepackage{algorithmic}
\usepackage[ruled]{algorithm2e}

\SetAlFnt{\small}
\SetAlCapFnt{\small}
\SetAlCapNameFnt{\small}
\SetAlCapHSkip{0pt}

\usepackage{mathtools}
\usepackage{balance}

\usepackage{float}
\usepackage{placeins}
\usepackage{multicol}
\usepackage{enumitem}
\usepackage{comment}
\usepackage{pifont}
\usepackage[toc,page]{appendix}

\usepackage{soul}
\usepackage{acronym}
\usepackage{fancybox}
\usepackage{epigraph}
\usepackage{dirtytalk}
\usepackage{bm}
\usepackage{fontawesome}
\usepackage{colortbl}
\usepackage{booktabs}

\usepackage{tikz}
\usepackage{listings}
\usepackage{microtype}
\usepackage{transparent}
\usepackage{hyphenat}
\usepackage{makecell}

\usepackage{needspace}

\crefname{section}{Sec.}{Secs.}
\Crefname{section}{Section}{Sections}
\Crefname{table}{Table}{Tables}
\crefname{table}{Tab.}{Tabs.}
\crefname{algocf}{Alg.}{Algs.}
\Crefname{algocf}{Algorithm}{Algorithms}

\usepackage[normalem]{ulem}
\usepackage{orcidlink}

\usepackage[flushleft]{threeparttable}

\usepackage{caption}

\definecolor{citecolor}{HTML}{0071bc}
\definecolor{frontcolor}{HTML}{325ea5}
\definecolor{backcolor}{HTML}{a58b77}
\definecolor{sidecolor}{HTML}{10768c}
\definecolor{skincolor}{HTML}{dcb7b7}
\definecolor{darkred}{rgb}{0.6, 0.1, 0.05}
\definecolor{DeltaColor}{rgb}{0.039,0.73,0.71}
\definecolor{SigmaColor}{rgb}{0.98,0.45,0.0}
\definecolor{AlphaColor}{rgb}{0,0,0.8}
\definecolor{BetaColor}{rgb}{0.8,0,0.8}
\definecolor{GammaColor}{rgb}{0.514,0.34,0.224}
\definecolor{EpsilonColor}{rgb}{0.353,0.725,0.906}
\definecolor{PurpleColor}{HTML}{9839ff}
\definecolor{BadColor}{HTML}{C0392B}
\definecolor{OrangeColor}{rgb}{0.914,0.541,0.0.141}
\definecolor{GreenColor}{HTML}{00ab41}
\definecolor{LightBlue}{HTML}{7dbaf3}
\definecolor{RedColor}{rgb}{0.949,0.275, 0.224}
\definecolor{LightCyan}{rgb}{0.88,1,1}
\definecolor{Gray}{gray}{0.85}
\definecolor{LightGray}{gray}{0.70}

\definecolor{greenprior}{HTML}{34a853}
\definecolor{redprior}{HTML}{ea4335}
\definecolor{blueprior}{HTML}{4285f4}

\definecolor{bestcolor}{rgb}{1, 0.5, 0.25}
\definecolor{secondbestcolor}{rgb}{1, 0.8, 0.5}

\newcommand{\bgood}[1]{\textcolor{GreenColor}{#1\%}}

\definecolor{bonecolor}{HTML}{34a853}
\definecolor{skelcolor}{HTML}{ea4335}

\newcommand{\skelebone}{\textcolor{skelcolor}{skele}\textcolor{bonecolor}{bones}\xspace}
\newcommand{\bones}{\textcolor{bonecolor}{bones}\xspace}

\newcommand{\ie}{\mbox{i.e.}\xspace}
\newcommand{\eg}{\mbox{e.g.}\xspace}

\newcolumntype{a}{>{\columncolor{Gray}}c}

\newcommand{\qheading}[1]{\noindent\textbf{#1.}}
\newcommand{\medheading}[1]{\medskip\noindent\textbf{#1.}}

\newlength\savewidth\newcommand\shline{\noalign{\global\savewidth\arrayrulewidth
  \global\arrayrulewidth 1pt}\hline\noalign{\global\arrayrulewidth\savewidth}}

\usepackage{amsmath}

\DeclareMathOperator*{\argmin}{arg\,min}

\usepackage{etoolbox}
\makeatletter
\AfterEndEnvironment{algorithm}{\let\@algcomment\relax}
\AtEndEnvironment{algorithm}{\kern2pt\hrule width 0.9\linewidth\vskip3pt\@algcomment}
\let\@algcomment\relax
\newcommand\algcomment[1]{\def\@algcomment{\footnotesize#1}}
\renewcommand\fs@ruled{\def\@fs@cfont{\bfseries}\let\@fs@capt\floatc@ruled
  \def\@fs@pre{\hrule height.8pt depth0pt \kern2pt}%
  \def\@fs@post{}%
  \def\@fs@mid{\kern2pt\hrule\kern2pt}%
  \let\@fs@iftopcapt\iftrue}
\makeatother

\newlist{SubItemList}{itemize}{1}
\setlist[SubItemList]{label={$-$}}

\let\OldItem\item
\newcommand{\SubItemStart}[1]{%
    \let\item\SubItemEnd
    \begin{SubItemList}[resume]%
        \OldItem #1%
}
\newcommand{\SubItemMiddle}[1]{%
    \OldItem #1%
}
\newcommand{\SubItemEnd}[1]{%
    \end{SubItemList}%
    \let\item\OldItem
    \let\SubItem\SubItemFirst
    \item #1%
}
\newcommand{\SubItemFirst}[1]{%
    \let\SubItem\SubItemMiddle%
    \SubItemStart{#1}%
}
\let\SubItem\SubItemFirst

\let\OldEndItemize\enditemize
\renewcommand{\enditemize}{%
    \ifx\item\SubItemEnd%
        \end{SubItemList}%
        \let\item\OldItem%
        \let\SubItem\SubItemFirst%
    \fi%
    \OldEndItemize%
}

\newcommand{\page}{\href{https://cookmaker.cn/gaussianimate/}{\textcolor{magenta}{\xspace\tt\textit{cookmaker.cn/gaussianimate}}\xspace}}

\newcommand{\suppl}{\textcolor{magenta}{\emph{Sup.Mat.}}\xspace}

\newcommand{\xmark}{\textcolor{RedColor}{\ding{55}}\xspace}
\newcommand{\cmark}{\textcolor{GreenColor}{\ding{51}}\xspace}

\newcommand{\dna}{\mbox{DNA-Rendering}\xspace}
\newcommand{\actor}{\mbox{ActorHQ}\xspace}

\newcommand{\modelname}{\textcolor{black}{\mbox{GaussiAnimate}}\xspace}
\newcommand{\modelnameLong}{Rig Animatable Categories with Level of Dynamics}
\newcommand{\ourtitle}{\textbf{\modelname}: \modelnameLong}

\acrodef{amt}[AMT]{Amazon Mechanical Turk}

\makeatletter
\@ifundefined{hyxmp@parse@acmart}{}{\let\hyxmp@parse@acmart\relax}
\makeatother

\AtBeginDocument{%
  }

\begin{document}

\title{\ourtitle}
\author{Jiaxin Wang}
\email{wangjiaxin@westlake.edu.cn}
\affiliation{%
  \institution{Westlake University}
  \city{Hangzhou}
  \country{China}
}

\author{Dongxin Lyu}
\email{lyudongxin@westlake.edu.cn}
\affiliation{%
  \institution{Westlake University}
  \city{Hangzhou}
  \country{China}
}

\author{Zeyu Cai}
\email{caizeyu010612@gmail.com}
\affiliation{%
  \institution{Westlake University}
  \city{Hangzhou}
  \country{China}
}
\affiliation{%
  \institution{Nanjing University}
  \city{Nanjing}
  \country{China}
}

\author{Zhiyang Dou}
\email{zhiyang0@connect.hku.hk}
\affiliation{%
  \institution{The University of Hong Kong}
  \city{Hong Kong}
  \country{China}
}

\author{Cheng Lin}
\email{chenglin@must.edu.mo}
\affiliation{%
  \institution{Macau University of Science and Technology}
  \city{Macau}
  \country{China}
}

\author{Anpei Chen}
\email{chenanpei@westlake.edu.cn}
\affiliation{%
  \institution{Westlake University}
  \city{Hangzhou}
  \country{China}
}

\author{Yuliang Xiu}
\authornote{Corresponding author.}
\email{xiuyuliang@westlake.edu.cn}
\affiliation{%
  \institution{Westlake University}
  \city{Hangzhou}
  \country{China}
}

\renewcommand{\shortauthors}{Wang et al.}

\begin{abstract}
Rigging dynamic 3D assets is essential for animated content creation, yet animating articulated objects with non-rigid surface dynamics—such as clothed humans, quadrupeds, and garments—demands reconciling two conflicting objectives: intuitive kinematic control and high-fidelity deformation modeling. Skeleton-based methods offer controllable articulation but struggle with complex surface dynamics, while free-form bone or blob-based representations capture non-rigid deformations but lack explicit kinematic structure.
%
%
%
To this end, we propose \textbf{Skelebones}, a Scaffold-Skin Rigging System built on three steps:
(1) \textit{Bones}: compress temporally-consistent Gaussian/Mesh sequences into free-form bones with smooth skinning weights, approximating non-rigid deformations via linear blend skinning (LBS);
(2) \textit{Skeleton}: extract the Mean Curvature Skeleton (MCS) from the canonical shape and temporally refine its topology and kinematics into a compact skeletal structure;
(3) \textit{Binding}: bind skeleton and bones via non-parametric Partwise Motion Matching (PartMM), which synthesizes novel bone motions by matching, retrieving, and blending existing ones.
Together, these steps compress the dynamics of 4D shapes into compact skelebones that are simultaneously controllable and expressive. The resulting representation is category-agnostic (\ie, template-free), motion-adaptive (\ie, dynamic topology), and topology-correct (\ie, skeleton consistent with surface geometry)—with PartMM requiring no learning.
We validate on both synthetic and real-world datasets, achieving substantial reanimation improvements on unseen poses: \bgood{17.3} dB PSNR gain over LBS at \dna and \bgood{45.6} dB over Bag-of-Bones (BoB) at \actor, while preserving high rendering fidelity for characters with complex non-rigid dynamics.
PartMM generalizes robustly to both Gaussian and Mesh representations, excelling in low-data regimes ($\sim$1000 frames) with a \bgood{48.4} RMSE improvement over LBS and outperforming GRU- and MLP-based methods by \bgood{>20}. Code will be publicly released in \page

{\small\noindent\textbf{Published version:}
\href{https://doi.org/10.1145/3829340.3842296}{doi:10.1145/3829340.3842296}.\par}

\keywords{Motion Matching \and Motion Retargeting \and Character Rigging and Skinning \and Gaussian-based 4D Reconstruction \and Non-rigid Deformation}
\end{abstract}

\begin{teaserfigure}
  \includegraphics[width=\textwidth]{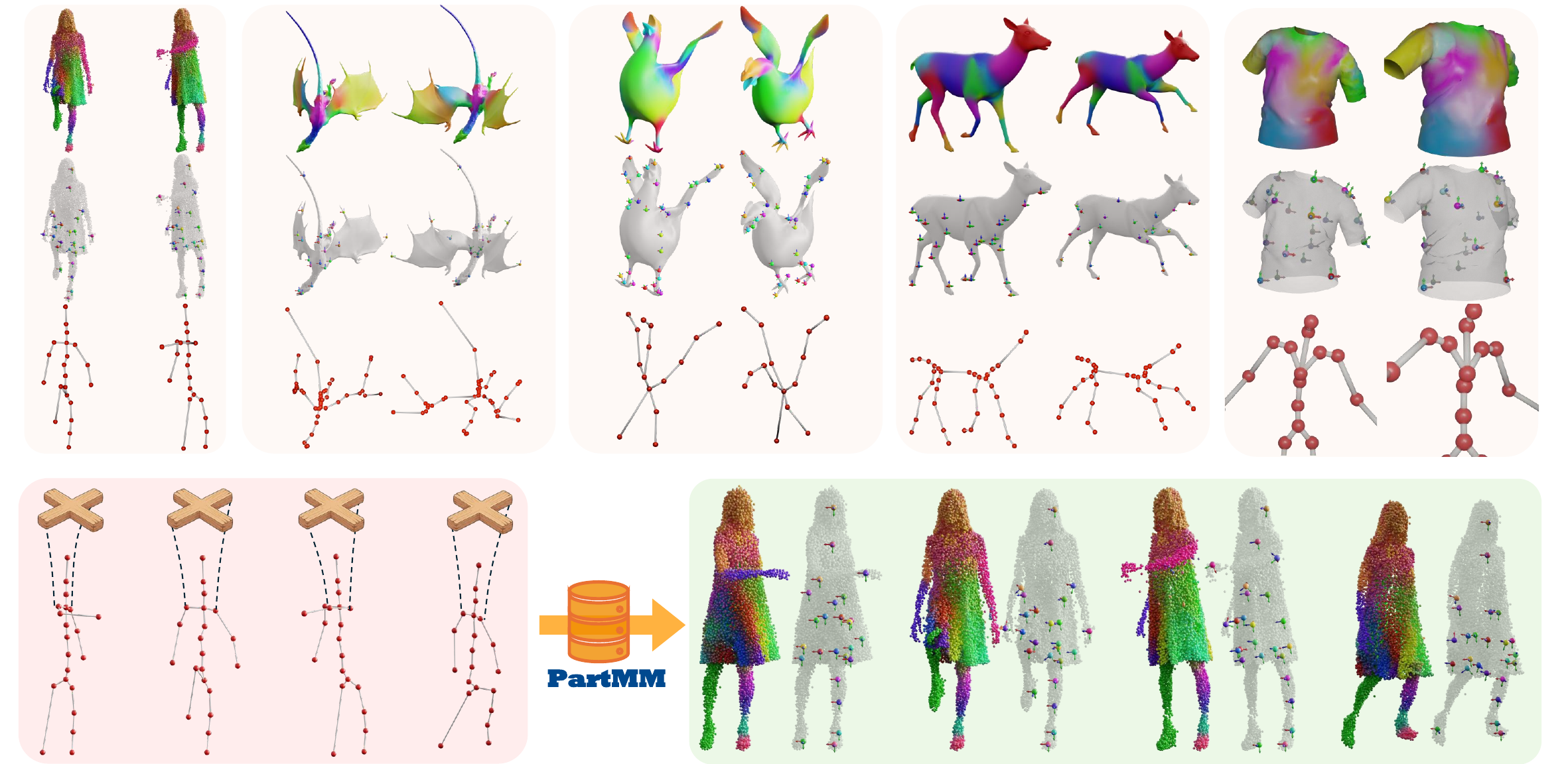}
  \caption{\textbf{\modelname} is designed to: 1) \textbf{rig} diverse animatable entities—typically featuring a soft exterior and rigid interior (\eg, clothed humans, quadrupeds, bipeds, birds, and garments)—from either reconstructed consistent 4DGS or mesh sequences. This relies on a novel \skelebone representation that balances the intuitive control of \textcolor{skelcolor}{kinematic skeletons} with the deformation fidelity of \textcolor{bonecolor}{free-form bones}; and 2) \textbf{animate} these rigged entities non-parametrically via Partwise Motion Matching (PartMM), in which skeletons drive the bones in a ``match-and-blend'' manner, enabling kinematic control of non-rigid deformations.
  }
  \label{fig:teaser}
\end{teaserfigure}

\maketitle
\hypersetup{%
  pdfcreator={LaTeX with pdfLaTeX},
  pdfpublisher={}%
}

\section{Introduction}
\label{sec:intro}

%
4D assets (sequential meshes or gaussians) are rapidly proliferating across domains: they are simulated by physics engines, captured by streaming sensors, reconstructed from video~\cite{dgmesh,pumarola2020dnerf,tan2025dressrecon}, and increasingly generated by recent generative AI methods~\cite{chen2026motion3to4,sabathier2026actionmeshanimated3dmesh}. As a result, they are poised to become foundational building blocks for immersive applications in gaming, VR, and film production, as well as for robotics digital twins that support sim-to-real learning.

However, raw 4D captures are not inherently animatable: they are recorded frame-by-frame for playback, not control. To unlock their value, we must convert and compress them into rigged assets that support controllable reanimation instead of passive replay. We focus on articulated categories with non-rigid surface deformations, including clothed humans, quadrupeds, bipeds, birds, and garments.


%

Rigging these categories presents an inherent \textit{``intuitive control vs. deformation fidelity''} trade-off. On one hand, the internal skeleton should be compact and rigid to enable intuitive control and efficient animation, such as with hierarchical skeleton~\cite{loper2023smpl,pavlakos2019expressive}. 
On the other hand, the outer surface must capture fine geometry details and complex non-rigid deformations (\eg, soft tissues, loose-fitting clothing), which demands a more expressive and flexible representation, such as virtual bones~\cite{SSDR,pan2022predicting}, bag of bones~\cite{tan2025dressrecon}, or free-form blobs~\cite{he2025category,epstein2022blobgan}. 

We address this controllability and fidelity trade-off via a scaffold-skin rigging system, termed \textbf{``\skelebone''}. We explicitly model the hierarchical \textit{Level of Dynamics} of the subject: it utilizes an inner kinematic \textcolor{skelcolor}{\textbf{skeleton}} to capture low-frequency rigid articulation, and outer free-form \textcolor{bonecolor}{\textbf{bones}} to represent high-frequency non-rigid deformations, with the latter driven by the former. Notably, both the skeletons and bones are directly extracted or optimized from the sequential shapes without relying on category-specific templates (\eg, SMPL~\cite{loper2023smpl}, SMAL~\cite{SMAL}) or strong priors, making our method broadly applicable across diverse categories.

This design enables intuitive control while preserving natural, complex deformations.
While this may seem straightforward in principle, in practice, it presents three main challenges: 
\textbf{1) Bones:} computing outer bones requires temporally consistent geometry reconstruction. How to maintain such consistency without compromising rendering quality is non-trivial; 
\textbf{2) Skeletonization:} extracting a kinematic skeleton that is topologically correct, category-agnostic (\ie, template-free), and motion-aware (\ie, able to adapt given more observations) is highly ill-posed, as suggested in \cref{tab:compare-s3o};
and \textbf{3) Binding:} propagating motion from the skeleton to the bones automatically, and generalizing to entirely unseen poses is highly under-constrained given limited video observations. 
To address above challenges, we make the following key technical choices: 

\medheading{1) Bones Compression} To address temporal inconsistency and geometric invalidity, we adopt a deformable 3DGS and enforce local rigidity via As-Rigid-As-Possible (ARAP)~\cite{ARAP}. Furthermore, we isolate rigid parts on surface via motion-guided clustering based on the \textit{Gestalt law of common fate}\footnote{\scriptsize The Gestalt law of common fate states that humans perceive visual elements that move in the same direction and at the same speed as being part of a single, related group.}. This structurally consistent initialization enables SSDR~\cite{SSDR} to compute free-form bones and skinning weights to model the non-rigid dynamics in a LBS system~\cite{lewis2023pose}.

%

\begin{table}
  \centering 
  \caption{Difference between existing methods. The \textbf{skeleton} means methods use a kinematic tree for animation and deformation, and the \textbf{bones} means methods use 6DOF free-form bones (also termed as blobs). \textcolor{orange}{Orange columns} are skeleton-related, while \textcolor{greenprior}{Green columns} are bone-related. }
  \label{tab:compare-s3o}
  \setlength{\tabcolsep}{2pt} 
  \resizebox{0.46\textwidth}{!}{
  \begin{threeparttable}
  \begin{tabular}{lc>{\columncolor{orange!30}}c >{\columncolor{orange!30}}c >{\columncolor{orange!30}}c >{\columncolor{greenprior!20}}c}
    Method &
    Rigging &
    \makecell{Template\\Free} &
    \makecell{Motion\\Adaptive} &
    \makecell{Topology\\Correct} &
    \makecell{Non-rigid\\Dynamic} \\
    \shline
    Robust~\cite{Robust}   & Skeleton & \cmark & \cmark & \xmark &  \xmark \\
    TAVA~\cite{li2022tava}      & Skeleton   & \xmark   & \xmark & \cmark & \xmark \\
    BANMo~\cite{yang2022banmo}      & Bones  & \cmark   & \cmark & \xmark &  \cmark \\
    RAC~\cite{yang2023rac}        & Skeleton & \xmark & \cmark & \cmark &  \xmark \\
    CAMM~\cite{kuai2023cammbuildingcategoryagnosticanimatable}      & Skeleton  & \cmark & \xmark & \xmark &  \xmark \\
    AP-NeRF~\cite{ap-nerf}     & Skeleton   & \cmark   & \xmark & \xmark & \xmark \\
    WIM~\cite{watchitmove}       & Skeleton   & \cmark & \cmark & \xmark &  \xmark \\
    SC-GS~\cite{scgs}   & Bones  & \cmark & \xmark & \xmark &  \cmark \\
    DressRecon~\cite{tan2025dressrecon} & SMPL,Bones & \xmark & \xmark & \cmark & \cmark \\
    GART~\cite{lei2023gartgaussianarticulatedtemplate} & SMPL,Bones & \xmark & \xmark & \cmark & \cmark \\
    RigGS~\cite{yao2025riggs}  & Skeleton & \cmark & \cmark & \xmark & \xmark \\
    CANOR~\cite{CANOR} & Bones & \cmark & \xmark & \xmark & \cmark \\
    \shline
    Ours        & Skelebones  & \cmark & \cmark & \cmark & \cmark\\
  \end{tabular}
  \begin{tablenotes}
      \item[\textbullet] \textbf{Template Free:} w/o predifined kinematic template (SMPL, SMAL, joints).
      \item[\textbullet] \textbf{Motion Adaptive:} Longer motion sequence $\rightarrow$ Better rigging.
      \item[\textbullet] \textbf{Topology Correct:} Temporally consistent kinematic skeleton.
      \item[\textbullet] \textbf{Non-rigid Dynamics:} Can model non-rigid deformation (clothing).
    \end{tablenotes}
  \end{threeparttable}}
\end{table}

\medheading{2) Skeleton Extraction} To address the skeletonization challenge, we first initialize a Mean Curvature Skeleton~\cite{tagliasacchi2012mean} directly from the canonical shape, removing the need for category-specific templates. We then locate joints at SSDR~\cite{SSDR} skinning-weight boundaries (\cref{fig:skeletonization}) to enforce topological correctness, and progressively refine the kinematic tree with extended motion observations to improve motion awareness. 


\medheading{3) Skelebones Binding} To propagate the inner motion to outer bones, we formulate bone binding as a non-parametric patch-based matching~\cite{chen2025motion2motion,granot2022drop,li2023example}. We treat cached skeleton poses as \textit{keys (K)}, queried novel skeleton poses as \textit{queries (Q)}, and cached non-rigid bone deformations as \textit{values (V)}. Crucially, short per-asset captured motion sequences often yield only small-scale skeleton-to-bone pairs. To overcome data scarcity, we introduce \textit{Partwise Motion Matching (PartMM)}, which matches and blends bone motions temporally (patch-level) and spatially (part-level) to produce plausible non-rigid motions for unseen poses.


Experiments on both synthetic and real-world datasets demonstrate the superiority of our approach. Specifically, \modelname achieves competitive novel-view synthesis quality and demonstrates compatibility with 4DGS pipelines while enabling animatability, see~\cref{tab:render_comp_synthetic}. Furthermore, as shown in~\cref{tab:render_comp_clothing}, PartMM shows remarkable generalization for novel-pose animation, yielding significant improvements over prior arts, with up to \bgood{17.3} dB PSNR over SMPL+LBS at \dna, and \bgood{45.6} dB over Bag-of-Bones (BoB) at \actor.
Particularly, in low-data regimes ($\sim$1000 frames) at~\cref{tab:animation_rmse_comp}, it delivers substantial quantitative leaps -- yielding up to \bgood{>20} RMSE decrease against neural-based methods (\ie, MLP, GRU) on VTO Tshirts, and \bgood{18.4} RMSE decrease over robust LBS on D4D animals.
%
Comprehensive ablations further validate each design choice, including motion adaptive (\cref{fig:progressive}), ARAP regularizer (\cref{fig:arap_ablation}), and partial vs. full skeleton matching (\cref{tab:animation_rmse_comp}). Also, we present qualitative results on various categories, including garments (\cref{fig:vto_comp}) and animals (\cref{fig:animal_comp}), demonstrating the versatility of \modelname.

Our main contributions are summarized as follows:
\begin{itemize}[leftmargin=*]
\item \textbf{Representation (Skelebones)}: a novel scaffold-skin rigging system that elegantly balances intuitive control with high-fidelity deformations by modeling the Level of Dynamics. It is template-free, motion-adaptive, and topology-correct.
\item \textbf{Algorithm (PartMM)}: a non-parametric, partwise matching algorithm that propagates kinematic motion to non-rigid surfaces, ensuring strong novel-pose generalization with limited data.
\end{itemize}

\section{Related Work}

\medheading{Shape- and Motion-aware Skeletonization}
Curve-skeleton and medial-axis methods derive compact structures directly from geometry~\cite{blum1967transformation,tagliasacchi2012mean,dey2006defining,tierny2007topology,reniers2008part,brunner2004mesh}. Because medial structures are sensitive to surface noise, later work uses pruning or volumetric primitives to improve robustness~\cite{QMAT,dou2022coverage,point2skeleton,wang2024coverage,miklos2010discrete}. Dynamic extensions such as D-MAT~\cite{DMAT} and Animated Sphere Mesh~\cite{ASM} fit medial primitives consistently across frames. Related sphere-mesh representations approximate and animate surfaces with spherical primitives~\cite{Thiery:2016:AMA:2903775.2898350,TGB:2013:SphereMesh}; in our representation, the outer spheres control surface motion while an inner kinematic skeleton provides the animation signal. We use a mean-curvature-flow curve skeleton~\cite{tagliasacchi2012mean} as the geometric scaffold, but locate joints from sequence-level SSDR skinning evidence~\cite{SSDR} rather than from the canonical shape alone.

\medheading{Feedforward and Per-asset Rigging}
Static-shape auto-rigging predicts joints, topology, and skinning weights from a single mesh, either with a prescribed structure~\cite{automaticrigging} or with learned models~\cite{song2025puppeteer,Song_2025_CVPR_magic_articulate,rignet,morig,liu2025riganything,zhang2025one,deng2025anymate,wang2026sprigselfsupervisedposeinvariantrigging,huang2026anigen,zhang2026skintokenslearnedcompact}. RigAnything~\cite{liu2025riganything}, for example, performs template-free feedforward rigging from one input shape. This differs from whether a method uses static or dynamic evidence: SSDR-style methods~\cite{SMA,SSDR,Example-based,Robust,CANOR} optimize free-form bones and skinning weights for each temporally consistent sequence, capturing its non-rigid deformation but producing controls without an anatomical kinematic hierarchy. Methods for 4D asset creation, such as ActionMesh~\cite{sabathier2026actionmeshanimated3dmesh} and Motion 3-to-4~\cite{chen2026motion3to4}, can provide dynamic geometry, but do not by themselves resolve the controllability of the resulting rig.\looseness=-1

RigMo~\cite{zhang2026rigmo} learns across raw mesh sequences without human-provided rig annotations and decodes Gaussian bones, skinning weights, and time-varying transformations for feedforward rig-and-motion inference. In contrast, \modelname takes a captured dynamic Gaussian sequence for one asset, uses no cross-asset training stage, and optimizes an explicit inner kinematic tree, outer SSDR bones, and an asset-specific motion database. Thus, RigMo targets scalable category-level generalization, whereas our experiments evaluate per-asset fitting, retrieval, and novel-pose reanimation. The two directions are complementary: per-asset optimization can provide high-fidelity rigging targets for future feedforward models.\looseness=-1

\begin{figure*}[t]
    \centering
    \includegraphics[width=\linewidth]{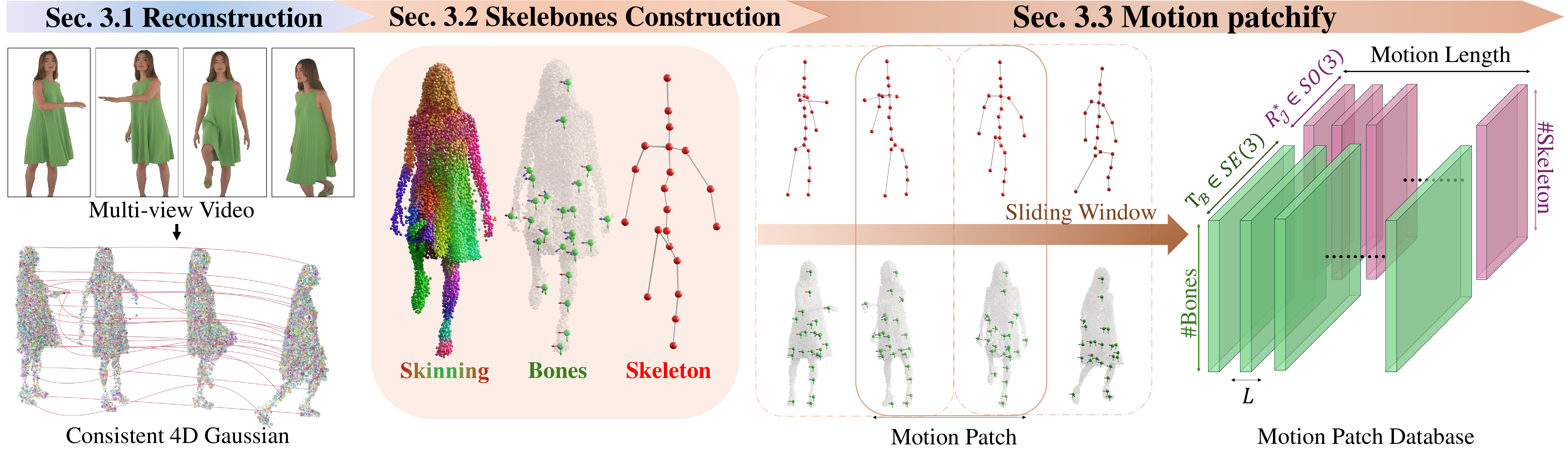}
    \caption{ \textbf{Pipeline Overview} Given a monocular or multi-view video, our method reconstructs a consistent 4DGS, extracts the inner skeleton via curve skeletonization~\cite{tagliasacchi2012mean} and the outer free-from bones via SSDR~\cite{SSDR}, together forming ``skelebones'', which are then used to build motion database. Point colors in the skinning and canonical-space views encode dominant outer-bone assignments.}
    \label{fig:pipe}
\end{figure*}

\medheading{Template-based and Template-free Animatable Reconstructions}
Neural representations, from Neural Radiance Fields~\cite{mildenhall2021nerf} to 3D Gaussian Splatting~\cite{3dgs}, support novel-view synthesis~\cite{chen2025human3r,chen2025feat2gs} and 3D reconstruction~\cite{cai2025up2you,chen2025easi3r,cai2024dreammapping,wang2025headevolver}. To make these representations animatable, template-based methods use predefined skeletons or parametric models such as SMPL~\cite{loper2023smpl} and SMAL~\cite{SMAL}, including RAC, TAVA, DressRecon, GART, ToMiE, and D3GA~\cite{yang2023rac,li2022tava,tan2025dressrecon,lei2023gartgaussianarticulatedtemplate,zhan2024tomiemodulargrowthenhanced,zielonka2024d3ga}. Their priors provide stable control but restrict object categories or motion spaces. Template-free methods instead discover parts or skeletons from reconstructions, including CAMM~\cite{kuai2023cammbuildingcategoryagnosticanimatable}, WIM~\cite{watchitmove}, AP-NeRF~\cite{ap-nerf}, SK-GS~\cite{SK-GS}, RigGS~\cite{yao2025riggs}, and BanMo~\cite{yang2022banmo}. These approaches expose a recurring trade-off: explicit skeletons provide controllability but can miss non-rigid surface motion, while free-form deformation models improve fidelity without necessarily yielding an intuitive kinematic rig. Our method is template-free and sequence-driven, and separates an explicit kinematic scaffold from free-form surface controls to retain both forms of motion.\looseness=-1

\medheading{Data-driven and Physics-based Deformation}
Physics-aware and simulation-ready avatar systems improve deformation beyond skeleton-only LBS via different mechanisms. PhysAvatar~\cite{zheng2024physavatar} estimates garment physics from multi-view video, while DiffAvatar~\cite{li2024diffavatar} recovers simulation-ready body and garment assets through differentiable simulation. Gaussian Garments~\cite{rong2025gaussiangarments} reconstructs simulation-ready clothing and adapts a neural simulator to observed garment behavior, and PhysRig~\cite{zhang2025physrigdifferentiablephysicsbasedskinning} embeds a skeleton in a simulated deformable volume. D3GA~\cite{zielonka2024d3ga} instead uses layered tetrahedral cages to capture volumetric effects such as garment sliding. These methods and ours address deformation fidelity through distinct routes: they introduce physical, simulation-ready, or cage-based deformation models, whereas PartMM reuses deformations observed in an asset-specific sequence. Our kinematics-driven formulation is category-agnostic and efficient at inference, but does not enforce thin-shell physics, collision constraints, or material parameters.

\section{Method}
\label{sec:method}

\paragraph{Input and scope.}
The method operates per asset. Its input is a temporally consistent sequence of Gaussian centers or mesh vertices; its output is an asset-specific skelebones rig and motion database. For video input, we first reconstruct the sequence with the deformable 3DGS model below. Thus, rigging is per-asset, not feedforward.


\subsection{Preliminaries}
\label{sec:preliminaries}

\qheading{Deformable 3DGS}
Vanilla 3DGS~\cite{3dgs} represents a static scene with numerous 3D Gaussians $\mathcal{G}$, where each Gaussian $g_i$ has attributes: position $\mu_i \in \mathbb{R}^3$, color $c_i \in \mathbb{R}^3$, opacity $\sigma_i \in \mathbb{R}$, rotation as quaternion $q_i \in \mathbb{R}^4$, and scale $s_i \in \mathbb{R}^3$.
Following RigGS~\cite{yao2025riggs}, we use isotropic Gaussians instead of anisotropic ones, trading reconstruction quality for better generalization to novel poses~\cite{lei2023gartgaussianarticulatedtemplate}—a necessary choice for reanimation.
Following SC-GS~\cite{scgs}, we represent dynamic scenes via a deformation field $D_\text{MLP}(f)$, parameterized by an MLP, that deforms the Canonical Gaussians $\mathcal{G}_c$ to frame $f$, denote as $\mathcal{G}_f$, across $F$ total frames. This deformation applies to all Gaussian attributes, denoted uniformly as $(\cdot)$:

\begin{equation}
    \mathcal{G}_{f}^{(\cdot)} = \mathcal{G}_c^{(\cdot)} + D_\text{MLP}^{(\cdot)}(f), \quad (\cdot) \in \{\mu,c,\sigma,q,s\},
\end{equation}

%
We optimize the deformable 3DGS using a photometric loss $\mathcal{L}_{\text{photo}}$ that combines L1 distance with a D-SSIM term, where $I_f$ and $\hat{I}_f$ denote the input and rendered images at frame $f$, respectively, and $\lambda_{\mathrm{ssim}}$ balances the two components. We use $\lambda_{\mathrm{ssim}}=0.5$ throughout:
\begin{equation}
    \mathcal{L}_{\text{photo}} = \sum_{f=1}^{F} \left[(1-\lambda_{\mathrm{ssim}}) \left\| I_f - \hat{I}_f \right\|_1 + \lambda_{\mathrm{ssim}} \left(1 - \text{D-SSIM}(I_f, \hat{I}_f)\right)\right].
\end{equation}

\qheading{Local Rigidity Regularization}
We regularize the Gaussians using ARAP term (As-Rigid-As-Possible)~\cite{ARAP}, and distance preservation term~\cite{2dgs} to encourage locally rigid deformations and the Gaussians to distribute on the surface:
\begin{equation}
\begin{aligned}
\mathcal{L}_{\text{rigid}}
&=
\lambda_{\mathrm{ARAP}}
\sum_{(i,j)\in K}
\underbrace{
\left\|
(x_f^i - x_f^j)
-
R_i (x_c^i - x_c^j)
\right\|_2^2
}_{\text{ARAP term}}
\\
&+
\lambda_{\mathrm{dist}}
\sum_{(i,j)\in K}
\underbrace{
\left(
\left\| x_f^i - x_f^j \right\|_2
-
\left\| x_c^i - x_c^j \right\|_2
\right)^2
}_{\text{distance preservation term}},
\end{aligned}
\label{eq:rigid_loss}
\end{equation}
where $K$ is the union of the 7-nearest-neighbor pairs constructed independently within each motion cluster. The number of motion clusters is capped at 50 by default. We set $\lambda_{\mathrm{ARAP}}=1$ and $\lambda_{\mathrm{dist}}=10$.
Here, $x_c^i$ and $x_f^i$ denote Gaussian $i$'s canonical and deformed positions, and $R_i \in \text{SO}(3)$ is obtained by Kabsch~\cite{Kabsch:a12999}; see \suppl for its derivation. The total objective is
\begin{equation}
\mathcal{L}_{\mathrm{recon}} = \mathcal{L}_{\mathrm{photo}} + \mathcal{L}_{\mathrm{rigid}}.
\end{equation}
Joint optimization of $\mathcal{L}_{\text{photo}}$ and $\mathcal{L}_{\text{rigid}}$ yields a time-consistent dynamic Gaussian sequence $G = \{\mathcal{G}_f\}_{f=1}^{F}$, where each frame deforms the canonical Gaussians $\mathcal{G}_c = \{g_i\}_\text{i=1}^N$ through the frame-dependent deformation field $D_{\text{MLP}}(f)$, and $N$ is the total number of Gaussians, which remains consistent across all frames.


\medheading{Motion Clustering}
After obtaining a time-consistent and locally rigid dynamic Gaussian sequence $G$, we further group Gaussian points into rigid clusters using a motion-aware clustering strategy following prior SSDR-based methods~\cite{SSDR}, producing coarse rigid transformations at each frame $f$, denoted as $\{\mathrm{T}_b^f\}_{b=1}^B$. Splitting stops when adding a cluster no longer decreases reconstruction error or when the maximum cluster count is reached. We set this maximum to 50 by default; the resulting cluster count defines the number $B$ of outer bones. See more details in \suppl's \cref{sup:motion_clustering}.\looseness=-1
%

\medheading{Skinning and Bones Decomposition}
Given the motion clusters and their corresponding rigid transformations $\mathbf{T}_{\hat{\mathcal{B}}} \in \text{SE(3)}^{F \times B}$, we employ the SSDR algorithm~\cite{SSDR} to fit Linear Blend Skinning (LBS) parameters. Specifically, SSDR computes the optimal skinning weights $\mathbf{W} \in \mathbb{R}^{N \times B}$ and refines the bone transformations to yield $\mathbf{T}_{\mathcal{B}} \in \text{SE(3)}^{F \times B}$, where $N$ and $B$ denote the number of Gaussians and rigid bones, respectively, consistent across all $F$ frames. This fitting compresses the dense 4D deformation into a sparse set of rigid bone transformations that approximate the surface motion, forming the \bones component of our \skelebone representation.
\begin{figure}[t]
    \centering
    \includegraphics[width=\linewidth]{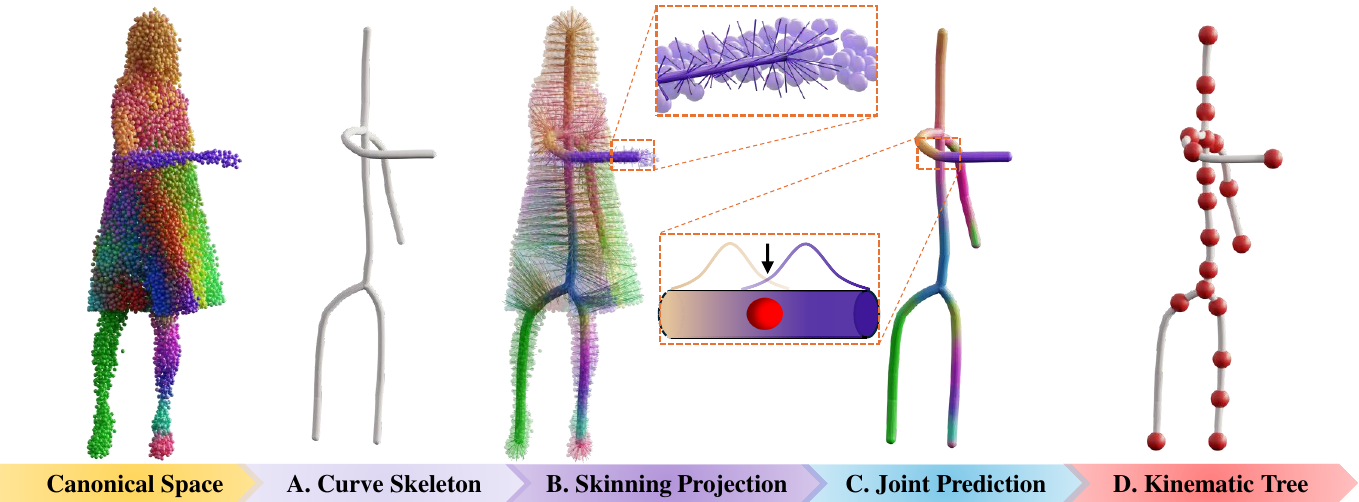}
    \caption{\textbf{Inner Skeleton Initialization.} We first extract the curve skeleton of the canonical shape to capture its structural connectivity (A). Next, the skinning weights of 3D points are projected onto the curve skeleton (B). We detect joint candidates by identifying sharp changes in skinning weights along the curve skeleton (C). Finally, we traverse the curve skeleton using Depth-First Search (DFS) to construct the kinematic tree of the skeleton (D).
    }
    \label{fig:skeletonization}
\end{figure}

\subsection{Inner Kinematic Skeleton Acquisition}
\label{sec:skeletonization}

While the bones obtained from SSDR effectively capture dominant surface deformations, they are essentially free-form and lack correspondence to anatomical structures, making intuitive and direct animation difficult. To address this, we extract the inner skeletal structure by performing curve skeletonization, identifying joint locations, constructing a kinematic tree, and optimizing skeletal poses via inverse kinematics. The overall pipeline is illustrated in \cref{fig:skeletonization}.

\medheading{Joints Selection from Curve Skeleton}
Given the canonical shape $\mathcal{G}_c$ (empirically chosen as frame 0), we first extract the Curve Skeleton~\cite{tagliasacchi2012mean} from $\mathcal{G}_c$, denoted as $\mathcal{S}$, which captures the underlying structural connectivity (see \cref{fig:skeletonization}-A). To identify joint locations, we leverage the observation that anatomical joints correspond to regions where the skinning weights $\mathbf{W}$ exhibit sharp spatial transitions~\cite{Robust,Example-based}. Therefore, we analyze the gradient of skinning weights along the curve skeleton to detect joint candidates (see \cref{fig:skeletonization}-C):
\begin{equation}
\mathcal{J} = \{ s_j \in \mathcal{S} \mid \nabla W(s_j) > \tau \}.
\label{eq:joint_detection}
\end{equation}
where $\mathcal{S} = \{ s_j \in \mathbb{R}^3 \mid j = 1, \ldots, \text{M} \}$ denotes the $\text{M}$ discrete set of canonical curve skeleton points, $\nabla W(s_j)$ denotes the gradient of skinning weights at skeleton point $s_j$, and we set $\tau=0.3$ by default to identify significant weight transitions. The projected weight at each skeleton point aggregates all associated surface points, so isolated projection errors do not directly create joints. These detected joints are expected to correspond to anatomical joint locations, providing a meaningful inner skeleton structure for animation control.

\medheading{Kinematic Tree Construction}
The detected joints are organized into a kinematic tree by establishing appropriate parent-child relationships. 
We construct a sparse skeleton representation $\{\mathcal{J}, \mathcal{E}\}$, where $\mathcal{E} \subset \mathcal{J} \times \mathcal{J}$ encodes the edges connecting detected joints, and $|\mathcal{J}| = \mathrm{J}$. 
Our strategy is as follows: (i) we retain edges that are spatially aligned with the curve skeleton, (ii) if the candidate graph contains a loop, we remove cycle edges before hierarchy construction using motion consistency over the sequence, and (iii) we determine parent-child relations using depth-first search (DFS) traversal, starting from the root joint, which is defined as the joint closest to the global center of mass. This process yields a kinematic tree $\mathcal{T} = (\mathcal{J}, \mathcal{A})$, as illustrated in~\cref{fig:skeletonization}-C, where $\mathcal{A}$ contains the parent index for each joint, establishing a hierarchical structure.
See further construction details in \suppl's \cref{sup:inner_skeleton}.

\medheading{Skeletal Pose Optimization}
Now we have obtained the kinematic skeleton in canonical space, we optimize the skeletal pose across frames by estimating local rotations for each joint $\mathbf{R}_{\mathcal{J}} = \{\mathbf{R}_j\}_{j=1}^\mathrm{J}$ with $\mathbf{R}_j \in \mathrm{SO}(3)$. We formulate this as an inverse kinematics problem that minimizes two objectives: (i) an L2 loss between Gaussians $\mathcal{G}_c$ deformed via forward kinematics with $\mathbf{R}_{\mathcal{J}}$ and the observed Gaussians $\mathcal{G}_f$ at frame $f$, and (ii) a Chamfer distance between the deformed skeleton joints $\mathcal{J}_c$ and the frame-wise curve skeleton $\mathcal{S}_f$:
\begin{equation}
\label{eq:ik_objective}
\mathbf{R}_{\mathcal{J}}^* =
\argmin_{\mathbf{R}_{\mathcal{J}}}
\sum_{f=1}^{F}
\left[
\mathcal{L}_{\mathrm{CD}}(\mathcal{J}_c(\mathbf{R}_{\mathcal{J}}), \mathcal{S}_f)
+
\mathcal{L}_{\mathrm{L2}}(\mathcal{G}_c(\mathbf{R}_{\mathcal{J}}), \mathcal{G}_f)
\right].
\end{equation}


\subsection{Partwise Motion Matching (PartMM)}
\label{sec:partmm}
\begin{figure*}[t]
    \centering
    \includegraphics[width=0.96\textwidth]{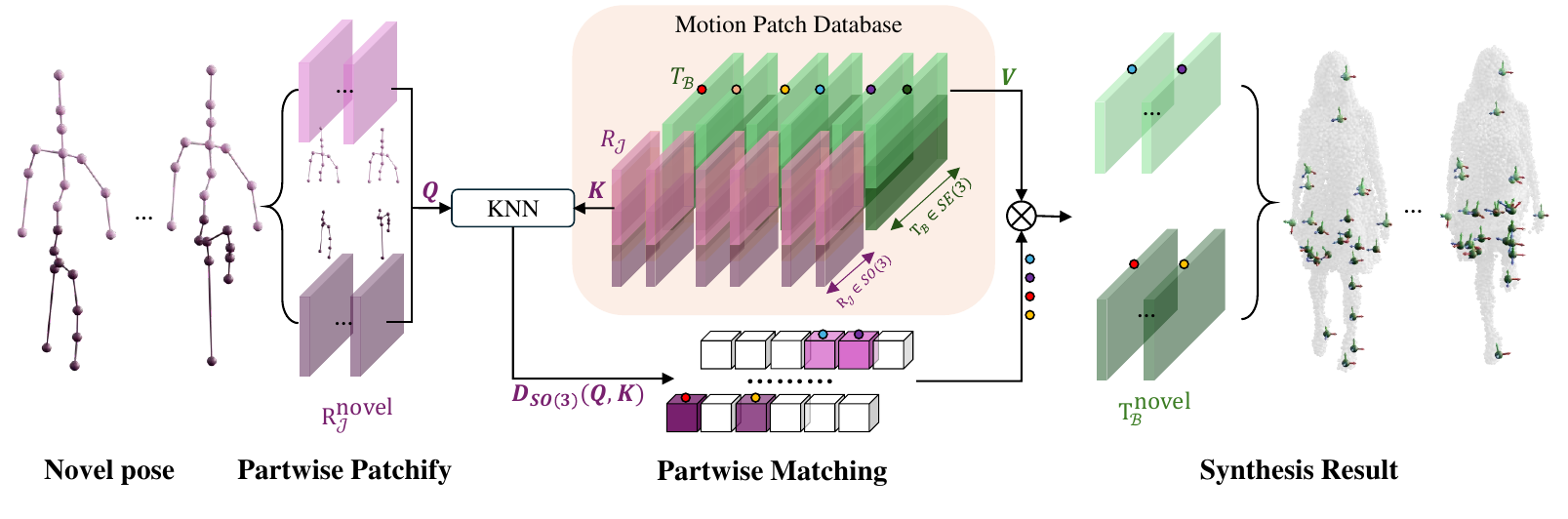}
    \caption{\textbf{Partwise Motion Matching (PartMM).} Given a novel inner-skeleton pose sequence, we animate skelebones by synthesizing outer-bone motion via part-wise matching. Our method: (a)~decomposes the kinematic tree into multiple parts (shown as two parts; user-defined in practice); (b)~extracts part-wise motion patches $R_{\mathcal{J}}^{\text{novel}}$ from the novel pose sequence; (c)~queries these patches against a pre-built motion database to retrieve similar patches to recompile, and then perform part-level spatial alignment. Iterating step~(c) yields the final motion $T_{\mathcal{B}}^{\text{novel}}$. Colored axes indicate each outer-bone control's local frame.}
    \label{fig:animation}
\end{figure*}

Now we have both the inner \textcolor{skelcolor}{joint rotations $\mathbf{R}_{\mathcal{J}} \in \mathrm{SO}(3)^{\mathrm{F \times J}}$}, \textcolor{bonecolor}{outer bone transformations $\mathbf{T}_{\mathcal{B}} \in \mathrm{SE(3)}^{F \times B}$}, and skinning weights $\mathbf{W} \in \mathbb{R}^{N \times B}$, forming our \skelebone representation. However, directly applying the observed bone transformations to animate novel poses is infeasible due to the lack of explicit correspondence between outer free-form bones and inner kinmatic skeleton. To address this, inspired by recent motion retargeting works~\cite{chen2025motion2motion, granot2022drop, li2023example}, we propose a partwise motion matching algorithm, shorten as PartMM, that synthesizes novel bone motions by retrieving and blending existing bone motion patches, guided by the inner skeletal motion. 
This process can be viewed as \textbf{KVQ (Key-Value-Query) retrieval system}: the target skeletal motion is the query (Q), source skeletal motions are keys (K), and source bone motions are values (V). Similarity is measured in $\mathrm{SO}(3)$ joint rotation space, and retrieved bone motions are blended for the final output. 
The pipeline is illustrated in \cref{fig:animation}, and \cref{alg:partmm} summarizes the complete matching, alignment, and blending procedure.
\begingroup
\setlength{\algomargin}{0pt}
\begin{algorithm}[!t]
\caption{Partwise Motion Matching, Alignment, and Blending.}
\label{alg:partmm}
\algcomment{\fontsize{7.2pt}{8pt}\selectfont \texttt{knn\_match}: $k$-nearest neighbors in $\mathrm{SO}(3)$; \\ \texttt{svd\_align}: optimal rigid alignment; \texttt{bmm}: batch matrix multiplication.}
\begingroup
\definecolor{codeblue}{rgb}{0.25,0.5,0.5}
\lstset{
  backgroundcolor=\color{white},
  basicstyle=\fontsize{7.2pt}{7.8pt}\ttfamily\selectfont,
  columns=fullflexible,
  breaklines=true,
  captionpos=b,
  commentstyle=\fontsize{7.2pt}{7.2pt}\color{codeblue},
  keywordstyle=\fontsize{7.2pt}{7.2pt}\color{black},
  xleftmargin=0pt,
}
\begin{lstlisting}[language=python]
# R_query: novel inner skeleton motion
# D_src: motion database
# parts: kinematic part groups

T_out = {}  # output outer bone transforms

for part in parts:
    # Query / Key / Value for this part
    Q = R_query[part]
    K = D_src[part].R
    V = D_src[part].T
    
    # kNN matching in SO(3) (Eq. 8)
    idx = knn_match(Q, K, k_neighbors=K_TOP)
    K_match = K[idx]
    V_match = V[idx]
    
    # rigid alignment via SVD (Eq. 9)
    R_align = svd_align(Q, K_match)
    V_aligned = bmm(R_align, V_match)
    
    # blend aligned outer motions
    T_out[part] = blend(V_aligned)

return T_out
\end{lstlisting}
\endgroup
\end{algorithm}
\endgroup

\medheading{Motion Patchifying}
We extract motion patches from the reconstructed skelebones sequence. Each frame contains:
(i) the local rotations of the inner skeleton joints, denoted as $\mathbf{R}_{\mathcal{J}} = \{\mathbf{R}_j\}_{j=1}^{\mathrm{J}}$ with $\mathbf{R}_j \in \mathrm{SO}(3)$, and
(ii) the 6-DoF rigid transformations of the outer bones, denoted as $\mathbf{T}_{\mathcal{B}} = \{\mathbf{T}_b\}_{b=1}^{B}$ with $\mathbf{T}_b \in \mathrm{SE}(3)$.
We group consecutive $p$ frames (default $p=7$) into motion patches and build a source motion database by sliding a temporal window:
\begin{equation}
\mathcal{D}_{\text{src}} = \{\mathcal{P}_l \mid \mathcal{P}_l = (\mathbf{R}_{\mathcal{J}}^l, \mathbf{T}_{\mathcal{B}}^l), \; l = 1, \ldots, L\}.
\end{equation}
where $L = F - p + 1$ is the number of patches.

\medheading{Partwise Patch Matching and Alignment}
Given a novel skeleton motion $\mathbf{R}_{\mathcal{J}}^{\text{novel}}$, we decompose it into five overlapping part-level motions according to the kinematic tree and use these as queries to retrieve relevant motion patches from the database. For each kinematic part $\mathcal{J}'$ (its corresponding bone part is $\mathcal{B}'$), we retrieve the $k=7$ nearest patches in the $\mathrm{SO}(3)$ space:
\begin{equation}
\mathcal{N}_k = \operatorname*{KNN}_{\mathcal{P}_l \in \mathcal{D}_{\text{src}}} \, d_{\mathrm{SO}(3)}(\mathbf{R}_{\mathcal{J}'}^{\text{novel}}, \mathbf{R}_{\mathcal{J}'}^{l}),
\end{equation}
\begin{figure}
    \centering
    \includegraphics[width=0.94\linewidth]{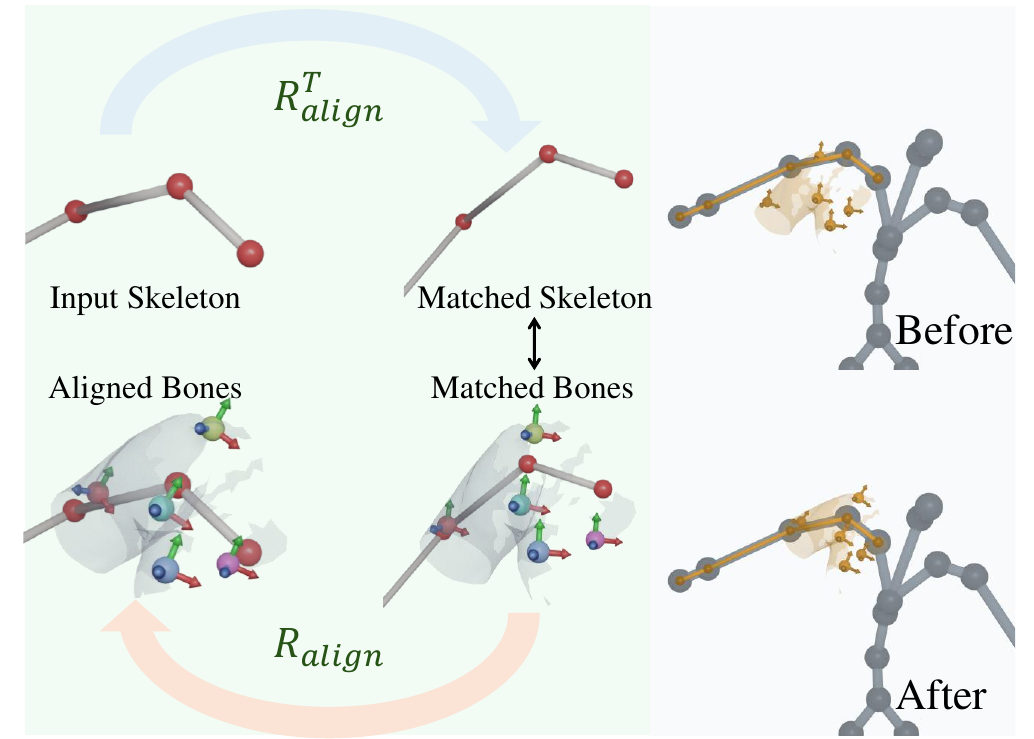}
    \caption{\textbf{Part Alignment}. Since perfect skeletal matching is rare, we further compute the optimal rotation to compensate.}
    \label{fig:part_to_whole}
\end{figure}

where $d_{\mathrm{SO}(3)}$ denotes the geodesic distance on the rotation manifold. PartMM uses no rejection threshold and always returns the nearest observed patches. Notably, patch matching is performed independently for each kinematic part, enabling flexible recombination of motion segments across parts, useful for synthesizing motions not directly present in the source data but composed from existing part-wise segments stored in the motion database.

To improve alignment between source and target skeletons, we compute the optimal rotation $\mathbf{R}_{\text{align}}$ via Kabsch~\cite{Kabsch:a12999} and apply it to bone transformations across each part, as shown in~\cref{fig:part_to_whole}:
\begin{equation}
\mathbf{T}_{\mathcal{B}'}^{\text{aligned}} = \mathbf{R}_{\text{align}} \circ \mathbf{T}_{\mathcal{B}'}^{\text{match}}.
\end{equation}

\medheading{Pyramid Refinement}
To further ensure temporal coherence and motion plausibility across the entire sequence, we additionally perform a refinement stage. We refine the retargeted bone motion using a motion pyramid by constructing multi-scale representations of the motion database. At each pyramid level $\ell$ (where $\ell=0$ is the coarsest), we downsample the motion sequence by a factor of $2^{5-\ell}$, creating progressively finer motion patches. Starting from coarse scales, we iteratively blend aligned motion patches:
\begin{equation}
\mathbf{T}_{\mathcal{B}'}^{\ell+1} = (1 - \lambda_\alpha) \, \mathbf{T}_{\mathcal{B}'}^{\ell} + \lambda_\alpha \, \bar{\mathbf{T}}_{\mathcal{B}'}^{\ell}.
\end{equation}
where $\bar{\mathbf{T}}_{\mathcal{B}}^{\ell}$ denotes the weighted average of matched patches at level $\ell$, and $\lambda_\alpha$ is the blending weight (default $\lambda_\alpha = 0.7$). We use five pyramid levels by default; see the pyramid-depth sensitivity analysis in \cref{sec:ablation}. This coarse-to-fine strategy enables smooth temporal transitions and geometric plausibility by progressively refining motion details across scales.
Finally, we apply forward skinning to synthesize novel poses with coherent local deformations across the surface under novel articulation.

\section{Experiments}
\subsection{Experimental Settings}
\label{sec:exp_setting}


\medheading{Datasets}

\textit{Synthetic datasets.} We use D-NeRF~\cite{pumarola2020dnerf} (8 sequences) and DG-Mesh~\cite{dgmesh} (6 sequences) to assess rendering quality on continuous action sequences. Following RigGS~\cite{yao2025riggs}, we exclude task-misaligned sequences---\textit{Bouncing Balls} (D-NeRF), \textit{Torus2Sphere} (DG-Mesh), and misaligned test views in \textit{Lego} (D-NeRF)---leaving 6 D-NeRF and 5 DG-Mesh sequences for all reported quantitative evaluations.

\textit{Real-world human data.} We evaluate our method on two clothed-human datasets with 8 subjects each from \dna~\cite{2023dnarendering} and \actor~\cite{isik2023humanrf}, split 80/20 for training and testing. Results validate animation of captured Gaussian assets under held-out poses across both evaluation datasets.

\textit{Non-rigid 4D meshes.} To validate generalization across different object categories and 3D representations, we conduct experiments on 4D garments via VTO~\cite{pan2022predicting} (109 motion sequences, 2 garment types, 13,979 cloth meshes) and 4D animals via D4D~\cite{li20214dcomplete} (59 animal types; we select 5 training and 1 testing sequence, $\sim$200 frames per animal). The VTO dataset includes SMPL skeleton as its inner skeleton, whereas the D4D dataset does not. We therefore preprocess the D4D dataset using our skeletonization pipeline to reconstruct skeletons for all sequences, using the output skeleton and optimized poses as the ground truth.
See \suppl (Secs.~F--G) for implementation and D4D preprocessing details.

\medheading{Baselines}
We compare PartMM against a comprehensive set of baselines for both reconstruction and animation.

\textit{For synthetic reconstruction} in~\cref{tab:render_comp_synthetic}, we include non-animatable 4D methods D-NeRF~\cite{pumarola2020dnerf}, TiNeuVox~\cite{TiNeuVox}, 4D-GS~\cite{Wu_2024_CVPR4dgs}, and SC-GS~\cite{scgs}, as well as animatable template-free methods AP-NeRF~\cite{ap-nerf} (NeRF with MAT) and RigGS~\cite{yao2025riggs} (a Gaussian approach that heuristically derives skeletons from SC-GS nodes~\cite{scgs}). Together, these baselines cover both reconstruction-only and template-free animatable methods in our evaluation.\looseness=-1

\textit{For Gaussian-based animation} in~\cref{tab:render_comp_clothing}, we evaluate Linear Blend Skinning (LBS)~\cite{abdrashitov2023robust,loper2023smpl}, Bag of Bones (BoB)~\cite{tan2025dressrecon}, and neural baselines (MLP~\cite{grigorev2023hood} and GRU~\cite{pan2022predicting}). All methods use Gaussians reconstructed by D3DGS~\cite{luiten2023dynamic}.\looseness=-1

\textit{For mesh animation} in~\cref{tab:animation_rmse_comp}, we compare Ours+LBS, Ours+MLP, and Ours+GRU. Specifically, Ours+LBS transfers clothing skinning weights from SMPL to simulated Tshirt or dress meshes~\cite{abdrashitov2023robust} and computes animal skinning weights with Bounded Biharmonic Weights (BBW)~\cite{BBW:2011}, while Ours+MLP and Ours+GRU use pose-conditioned networks to predict bone motion~\cite{grigorev2023hood,pan2022predicting}.


\begin{table}[t]
\centering
\caption{\textbf{Unseen Views}. Comparisons of the average precision (PSNR $\uparrow$ / SSIM $\uparrow$ / LPIPS $\downarrow$) on the D-NeRF~\cite{pumarola2020dnerf} dataset and DG-Mesh~\cite{dgmesh} dataset in unseen views.}
\label{tab:render_comp_synthetic}
\resizebox{\linewidth}{!}{
\begin{tabular}{lc|cccccc|c}
\toprule
\multirow{2}{*}{Method} & \multirow{2}{*}{w/ Rig} 
& \multicolumn{3}{c}{\textbf{D-NeRF}}
& \multicolumn{3}{c|}{\textbf{DG-Mesh}}
& \multirow{2}{*}{Time} \\
& & PSNR$\uparrow$ & SSIM$\uparrow$ & LPIPS $\downarrow$
& PSNR$\uparrow$ & SSIM$\uparrow$ & LPIPS $\downarrow$
&\\

\midrule
D-NeRF~\cite{pumarola2020dnerf} & \xmark    & 30.48 & 0.973 & 0.0492 & 28.17 & 0.957 & 0.0778 & $\textgreater$1day \\
TiNeuVox~\cite{TiNeuVox} & \xmark & 32.60 & 0.983 & 0.0436 & 31.95 & 0.967 & 0.0477 & $\approx$1hrs\\
4D-GS~\cite{Wu_2024_CVPR4dgs} & \xmark     & 33.25 & 0.989 & 0.0233 & 33.96 & 0.979 & 0.0272 & $\approx$0.5hrs \\
SC-GS~\cite{scgs} & \xmark     & \textbf{43.04} & \textbf{0.998} & \textbf{0.0066} & \textbf{38.96} & \textbf{0.993} & \textbf{0.0136} & $\approx$1hrs \\
\midrule
AP-NeRF~\cite{ap-nerf} & \cmark & 30.94 & 0.970 & 0.0350 & 31.83 & 0.967 & 0.0460 & $\approx$2hrs \\
RigGS~\cite{yao2025riggs} & \cmark       & 40.82 & \textbf{0.996} & \textbf{0.0112} & \textbf{37.65} & \textbf{0.991} & \textbf{0.0169} & $\approx$2hrs \\
Ours & \cmark & \textbf{41.00} & \textbf{0.996} & \underline{0.0154} & \underline{37.59} & \underline{0.990} & \underline{0.0220} & \textbf{$\approx$15min}\\
\bottomrule
\end{tabular}
}
\end{table}








\begin{table}[t]
\centering
\caption{
\textbf{Unseen Poses for Clothing.}
Comparisons of the rendering quality on the \dna~\cite{2023dnarendering} and \actor~\cite{isik2023humanrf} in unseen poses. The D3DGS refers to the multi-view reconstruction result as GT, SMPL+LBS refers to SkinTransfer~\cite{abdrashitov2023robust}, and BoB means the ``Bag of Bones'' introduced in DressRecon~\cite{tan2025dressrecon}.
}
\label{tab:render_comp_clothing}

\resizebox{\linewidth}{!}{
\begin{tabular}{lcccccc}
\toprule
\multirow{2}{*}{Method}
& \multicolumn{3}{c}{\textbf{\dna}}
& \multicolumn{3}{c}{\textbf{ActorsHQ}} \\
& PSNR$\uparrow$
& SSIM$\uparrow$
& LPIPS$\downarrow$
& PSNR$\uparrow$
& SSIM$\uparrow$
& LPIPS$\downarrow$ \\
\midrule

D3DGS(GT)~\cite{luiten2023dynamic}
& 32.16 & 0.978 & 0.033
& 30.56 & 0.938 & 0.136 \\

\midrule

SMPL+LBS~\cite{abdrashitov2023robust}
& 24.10 & 0.948 & 0.049
& 17.54 & 0.738 & 0.315 \\

SMPL+BoB~\cite{tan2025dressrecon}
& 23.03 & 0.935 & 0.057
& 16.66 & 0.722 & 0.259 \\

SMPL+PartMM
& 28.28 & 0.964 & 0.039
& 24.26 & 0.850 & 0.182 \\
\bottomrule
\end{tabular}
}
\end{table}

\begin{table}[t]
\centering
\caption{
\textbf{Unseen Poses for Meshes.}
RMSE comparisons of Robust LBS~\cite{abdrashitov2023robust}, neural variants (\emph{Ours$_\mathrm{MLP}$}, \emph{Ours$_\mathrm{GRU}$}), and motion-matching variants (\emph{FullMM}, \emph{PartMM}) on VTO and D4D.
}
\label{tab:animation_rmse_comp}
\resizebox{\linewidth}{!}{
\begin{tabular}{lccc}
\toprule
\multirow{2}{*}{Method}
& \textbf{VTO Tshirt}
& \textbf{VTO Dress}
& \textbf{D4D Animals} \\
& RMSE$\downarrow$
& RMSE$\downarrow$
& RMSE$\downarrow$ \\
\midrule
Robust LBS~\cite{abdrashitov2023robust} & 33.7 & 52.2 & 65.3 \\
Ours$_\mathrm{MLP}$ & 23.9 & 36.8 & 68.8 \\
Ours$_\mathrm{GRU}$ & 22.9 & 37.5 & 58.4 \\
FullMM & 38.2 & 37.0 & \textbf{53.3} \\
PartMM & \textbf{17.4} & \textbf{36.7} & 54.6 \\
\bottomrule
\end{tabular}
}
\end{table}

\subsection{Quantitative Comparison}

\medheading{Novel View Synthesis}
\Cref{tab:render_comp_synthetic} compares novel-view synthesis on D-NeRF~\cite{pumarola2020dnerf} and DG-Mesh~\cite{dgmesh} datasets against both non-animatable and animatable baselines. Our method achieves slightly worse rendering quality than non-animatable baselines, which is expected, as the additional consistency regularizers introduced onto SC-GS~\cite{scgs} trade off some rendering fidelity for animatability. However, it achieves competitive rendering quality to RigGS~\cite{yao2025riggs} with substantially faster optimization time ($\approx$15 minutes vs. $\approx$2 hours).

This gain stems from decoupling rigging from rendering-in-the-loop optimization. RigGS tightly couples skeleton extraction and skinning to rendering, requiring two photometric stages: 80,000 iterations to initialize ``skeleton-aware nodes'', followed by another 100,000 iterations to train MLPs that predict skinning, skeletal motion, and residual deformation. In contrast, \modelname decouples rigging from photometric optimization. Once a temporally consistent 4D Gaussian sequence is reconstructed, skelebones construction---including SSDR-based bone decomposition, skeleton extraction, and skinning-weight computation---runs directly in 3D without further photometric backpropagation.\looseness=-1


\medheading{Novel Pose Animation}
\Cref{tab:render_comp_clothing,tab:animation_rmse_comp} evaluate reanimation under unseen poses from both rendering and geometric perspectives.
LBS misses complex clothing dynamics, while neural binding degrades under pose extrapolation. Our method improves PSNR over SMPL+LBS on \dna and SMPL+BoB on \actor without pose-conditioned retraining (\cref{tab:render_comp_clothing}). On VTO, PartMM reduces RMSE relative to robust LBS, GRU, and MLP (\cref{tab:animation_rmse_comp}), supporting part-level recombination in low-data regimes. On D4D, both matching variants outperform neural and traditional baselines; FullMM is slightly better because the animal motions are highly repetitive.\looseness=-1


%
%
\begin{figure*}[t!]
    \centering
    \begin{minipage}{\textwidth}
        \centering
        \includegraphics[width=\linewidth,keepaspectratio]{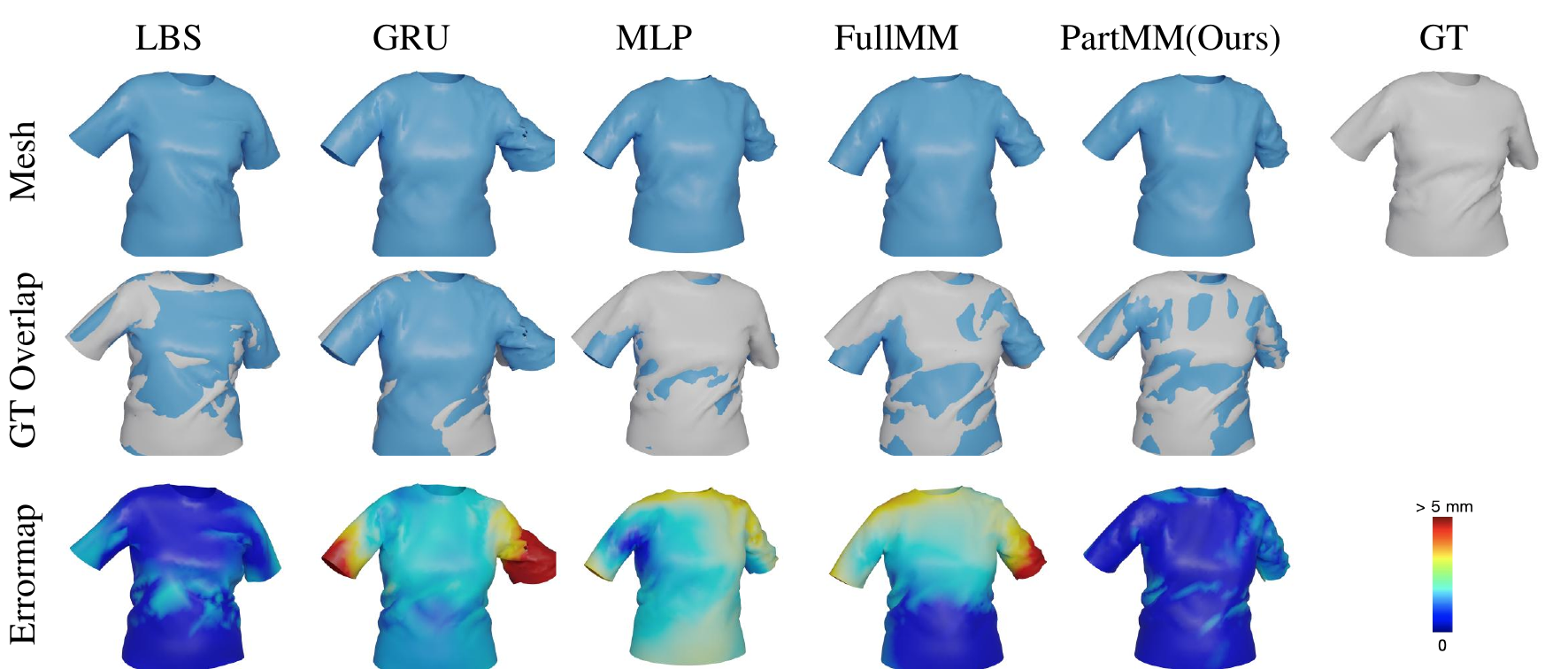}
        \captionof{figure}{\textbf{Qualitative Comparison on VTO dataset.}
        Reconstructed meshes, error maps, and ground-truth overlaps (top to bottom). PartMM yields lower errors than FullMM and generalizes better to unseen motions.}
        \label{fig:vto_comp}
    \end{minipage}
    \par\medskip
    \begin{minipage}{\textwidth}
        \centering
        \includegraphics[width=\linewidth,keepaspectratio,trim=0 7mm 0 0,clip]{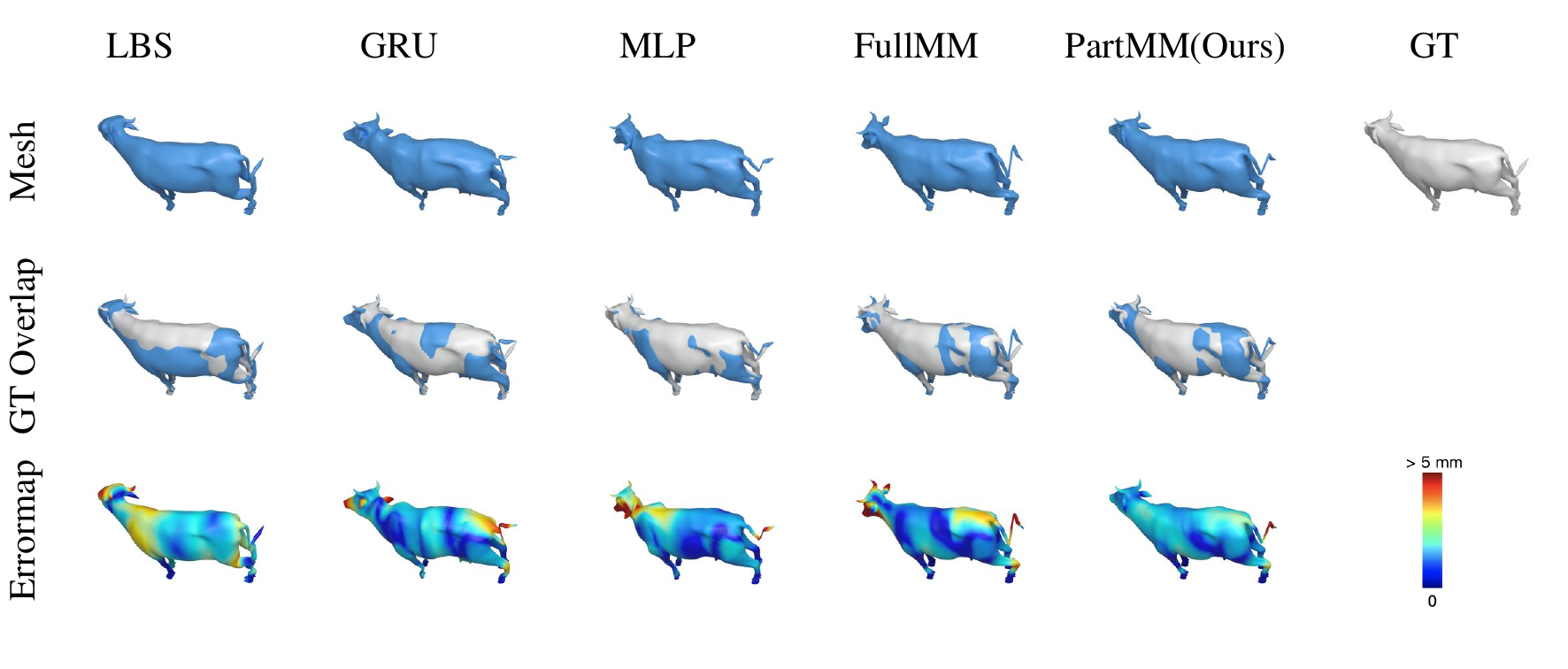}
        \captionof{figure}{\textbf{Qualitative Comparison on D4D dataset.}
        Reconstructed meshes, error maps, and ground-truth overlaps show that PartMM captures complex skinned animal deformation across diverse categories.}
        \label{fig:animal_comp}
    \end{minipage}
\end{figure*}

\subsection{Qualitative Comparison}

\renewcommand{\dbltopfraction}{0.95}
\renewcommand{\textfraction}{0.05}
\setcounter{dbltopnumber}{2}

\Cref{fig:vto_comp} demonstrates more accurate and natural deformations under unseen poses compared to classic LBS and data-driven baselines (MLP/GRU) on the VTO dataset.
\Cref{fig:animal_comp} shows that PartMM robustly captures complex skinned deformations across diverse animal categories. Overall, PartMM offers a unified solution for animating assets with a soft exterior and rigid core~(\eg, clothed humans, quadrupeds, bipeds, birds, and garments).

\subsection{Ablation Study}
\label{sec:ablation}

We assess how progressive skeletonization, ARAP regularization, and SSDR-based skinning affect skeletal structure, motion clustering, and animation stability throughout the pipeline.

\begin{figure*}[t!]
    \centering
    \begin{minipage}[t]{0.26\textwidth}
        \centering
        \includegraphics[width=\textwidth]{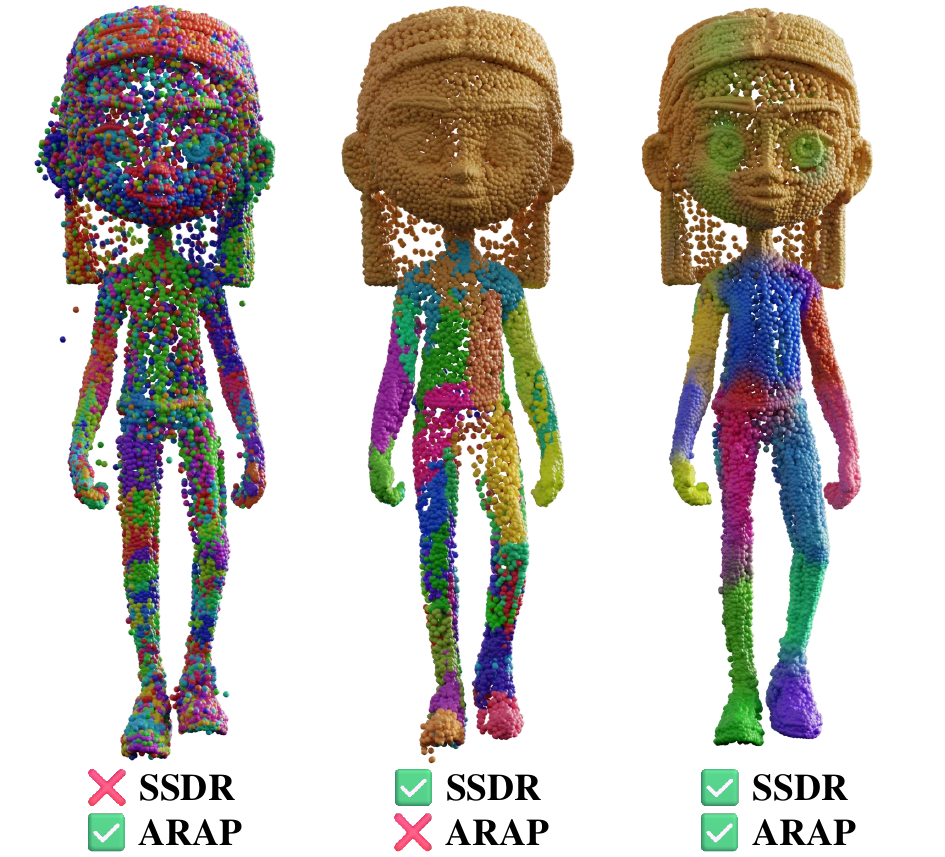}
        \captionof{figure}{\textbf{ARAP ablation.}}
        \label{fig:arap_ablation}
    \end{minipage}
    \hfill
    \begin{minipage}[t]{0.72\textwidth}
        \centering
        \includegraphics[width=\textwidth]{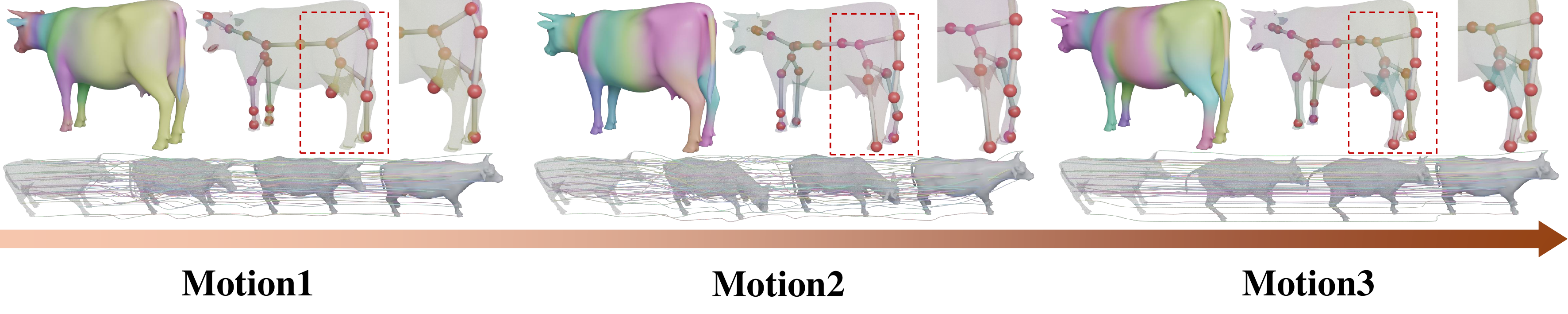}
        \captionof{figure}{\textbf{Progressive skeletonization.} More frames yield a more plausible, better-aligned skeleton.}
        \label{fig:progressive}
    \end{minipage}
\end{figure*}

\noindent\textbf{Progressive skeletonization.}
\Cref{fig:progressive} shows that additional motion observations expose more poses and deformations, progressively improving motion clustering, skinning estimation, and recovery of fine skeletal structure over successive frames.

\noindent\textbf{ARAP and SSDR.}
ARAP stabilizes motion clustering by suppressing local deformation (\cref{fig:arap_ablation}). Without SSDR's smooth decomposition, an over-dense bone set~\cite{yao2025riggs,scgs} produces fragmented skinning and unstable skeletonization.\looseness=-1

\noindent\textbf{PartMM patch length.}
We vary the temporal patch length $p$ on VTO while keeping all other settings fixed. Short patches remain stable, while long windows are harder to match in a limited database (\cref{tab:partmm_patch_length}). We retain the submitted $p=7$ because it provides longer temporal context, despite the slightly lower error at $p=5$.

\begin{table}[!ht]
    \centering
    \caption{Effect of PartMM patch length on VTO; the submitted setting is marked by an asterisk, and the lowest RMSE is shown in bold.}
    \label{tab:partmm_patch_length}
    \setlength{\tabcolsep}{5.2pt}
    \begin{tabular}{lccccc}
    \toprule
    $p$ (frames) & 3 & 5 & 7$^*$ & 11 & 15 \\
    \midrule
    RMSE (mm) & 17.17 & \textbf{16.80} & 17.36 & 18.79 & 20.56 \\
    \bottomrule
    \end{tabular}
\end{table}

\begin{samepage}
\noindent\textbf{Pyramid depth and blending.}
Increasing the pyramid depth from one to seven levels keeps RMSE within 17.18--17.63~mm while runtime rises from 18.6 to 133~ms per frame; we therefore use five levels by default to balance accuracy and cost. With $\lambda_\alpha=0$, no matched update is applied and the result remains at the initial pose (218~mm RMSE). Every tested $\lambda_\alpha>0$ converges to approximately 17.7~mm given enough refinement steps, indicating that the blending weight controls convergence speed rather than the converged solution.
\end{samepage}

\noindent\textbf{Database coverage.}
We use the minimum geodesic $\mathrm{SO}(3)$ distance from a query pose to the database as a coverage indicator. Larger databases improve accuracy and coverage, with diminishing returns beyond about 20 sequences, while build time, prediction time, and memory increase; \cref{tab:database_coverage} reports this accuracy--cost trade-off. Outside the observed coverage, PartMM retrieves the nearest patch and approaches an ARAP-like locally rigid approximation rather than physically faithful extrapolation beyond the database support.

\begin{table}[!ht]
    \centering
    \caption{PartMM accuracy, coverage, and cost versus database size on VTO.}
    \label{tab:database_coverage}
    \resizebox{\columnwidth}{!}{%
    \begin{tabular}{lccccc}
    \toprule
    Sequences & RMSE (mm) & min-$\mathrm{SO}(3)$ & Build (s) & Predict (ms) & GPU (MB) \\
    \midrule
    5  & 17.36 & 0.0988 & 10.5 & 16.0 & 144 \\
    20 & 15.74 & 0.0888 & 25.9 & 22.3 & 356 \\
    50 & \textbf{15.27} & 0.0736 & 66.3 & 27.0 & 1001 \\
    \bottomrule
    \end{tabular}}
\end{table}

\noindent\textbf{Skeleton robustness.}
We vary the joint threshold $\tau$ from 0.2 to 0.4 on 20 D4D cases. Eighteen cases retain the same joint count; two change by one joint, and the final RMSE changes by approximately 1 mm. Each curve-skeleton sample aggregates weights from roughly 1,000 surface projections, so isolated nearest-neighbor errors are averaged rather than forming spurious joints.

\noindent\textbf{Canonical-frame sensitivity.}
We compare the first, middle, last, and largest-bounding-box frames as the canonical frame on 20 D4D cases. Four cases change by one inner-skeleton joint, thirteen change by one to three outer bones, and the final RMSE varies by approximately 1 mm. The method is not tied to frame 0; if it contains strong self-contact, we select another frame with less self-contact.

%
%
%

\section{Conclusion}

In summary, we present \textbf{\modelname}, a unified framework that seamlessly combines deformable 3D Gaussians with a novel scaffold-skin rigging system. This design decouples two complementary roles: a kinematic skeleton governs global articulation, while free-form bones capture local non-rigid deformations. The two layers are integrated through partwise motion matching.\looseness=-1

Our approach offers an alternative to traditional rigging pipelines \cite{automaticrigging,rignet} and recent template-free approaches \cite{yao2025riggs,deng2025anymate} that rely on a single kinematic skeleton. It delivers intuitive control and natural deformations while requiring significantly lower computational overhead than simulation-based methods \cite{liu2013fast,li2020cipc,huang2024differentiable,wang2021gpu}. The framework is template-free and exhibits a key advantage of scalability: the quality of both the skelebones and reanimation improves progressively as more observations become available. In essence, richer motion capture enables better control and more faithful reanimation across diverse assets.

\medheading{Limitations and Future Work}
While our method demonstrates promising results, several avenues remain for improvement. Current computational performance is not yet real-time; however, several components present clear optimization opportunities. Database construction could be accelerated through neural compression~\cite{learnedMM} or neural SSDR~\cite{zhang2026rigmo}. Skeleton extraction remains sensitive to skinning quality and canonical-shape accuracy, especially for complex garments such as skirts, where estimated hip joints may deviate from anatomical expectations. Temporally consistent rigidity analysis with animated sphere meshes as volumetric priors offers a promising direction for more reliable skeleton extraction.\looseness=-1

Rendering artifacts under novel poses---such as holes in Gaussian surfaces caused by insufficient coverage---could be addressed by integrating mesh-based surface rendering or image-space refinement, as demonstrated in Animatable Gaussians~\cite{li2024animatablegaussians}. The quality of the Mean Curvature Skeleton also depends on the completeness of the canonical shape, which can be difficult to reconstruct from limited views. Learning-based extraction or temporal consistency across frames may further improve this stage.\looseness=-1

\textbf{Physical behavior and input coverage.} \modelname is a kinematics-driven rigging method rather than a cloth or soft-body simulator: outer-bone LBS does not enforce thin-shell mechanics or collision constraints and may therefore produce frozen wrinkles, miss secondary cloth motion, or permit body--garment penetration. PartMM also depends on coverage of the asset-specific motion database; outside that coverage, the retrieved deformation approaches a locally rigid approximation rather than physically valid extrapolation. The pipeline assumes temporally corresponded, locally rigid input motion, and imperfect foreground masks can introduce appearance artifacts. Physics-aware deformation, collision handling, and appearance refinement are important future directions.\looseness=-1





\medheading{Acknowledgments} 
We thank \textit{Ling-Hao Chen} for fruitful discussions on motion matching for retargeting~\cite{chen2025motion2motion}, which inspired our shift from learning-based to matching-based methods; \textit{Peizhuo Li} and \textit{Genshan Yang} for insightful feedback during the literature survey; \textit{Siyuan Yu} for testing and visualization support, as well as for dedicating substantial time and effort to producing a polished and visually compelling demo video; \textit{Yue Chen} and \textit{Xingyu Chen} for helpful suggestions on figure design, with \textit{Yue Chen} also providing meticulous proofreading of the manuscript; the members of \textit{Endless AI Lab} for their discussions and proofreading; and the \textit{ActorsHQ}, \textit{DNA-Rendering}, \textit{DeformingThings4D}, and \textit{VTO} dataset teams for providing the datasets. This work is supported by the Research Center for Industries of the Future (RCIF) at Westlake University, the Westlake Education Foundation, and the Macao Science and Technology Development Fund (FDCT) (0119/2025/ITP2).\looseness=-1

\bibliographystyle{ACM-Reference-Format}
\bibliography{reference}

\clearpage
\appendix

\renewcommand{\thefigure}{R.\arabic{figure}}
\renewcommand{\thetable}{R.\arabic{table}}
\renewcommand{\theequation}{S.\arabic{equation}}

\setcounter{figure}{0}
\setcounter{table}{0}
\setcounter{equation}{0}



\begin{subappendices}
\renewcommand{\thesection}{\Alph{section}}%
\label{sec:appendix}
\begin{figure*}[h!]
    \centering
    \includegraphics[width=\textwidth]{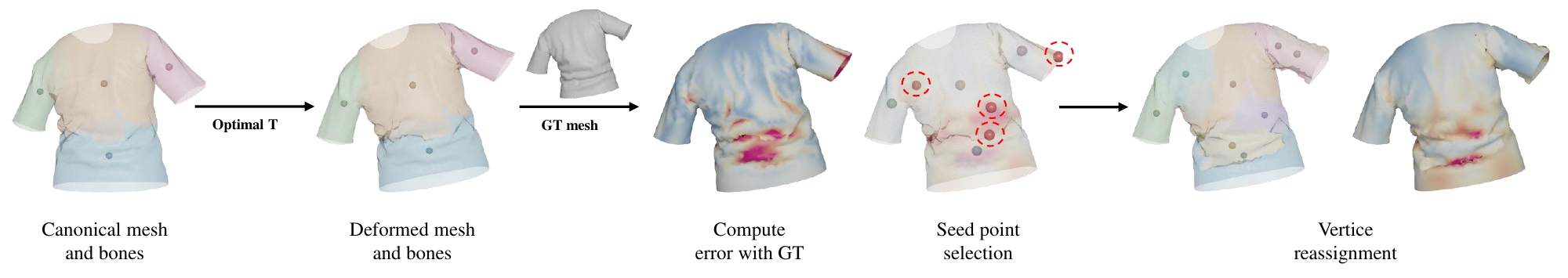}
    \caption{ \textbf{Spliting process of motion clustering}. We first compute the optimal transformation matrix $\mathbf{T}$ for the canonical mesh with a fixed number of bones (a). We then deform the mesh (b) and compute the reconstruction error against the ground truth, producing an error map (c). Based on this error map, we select the seed point corresponding to the largest reconstruction error (d). Finally, we recompute the reconstruction error using the transformation matrix of the newly added seed bone, and reassign each vertex to the seed point that minimizes its reconstruction error, thereby further reducing the overall error.}
    \label{fig:motion cluster}
\end{figure*}

\section{Motion Clustering}
\label{sup:motion_clustering}

We recursively cluster Gaussian motion into rigid parts.

\paragraph{Reconstruction Error.}
Let $\mathcal{G}_c = \{\mathbf{g}_i^c\}$ and $\mathcal{G}_f = \{\mathbf{g}_i^f\}$ denote the Gaussians in the rest pose and frame $f$, respectively. For each cluster $j$, rigid transformations $\{\mathbf{R}_j^f,\mathbf{T}_j^f\}$ are estimated using the Kabsch algorithm~\cite{Kabsch:a12999} from temporal correspondence established by the deformation function $D_\text{MLP}$ in \cref{sec:preliminaries}. The reconstruction error for point $i$ is defined as:
\begin{equation}
e_i = \sqrt{\sum_f 
\left\|
\mathbf{R}_{\ell(i)}^{f}\mathbf{g}_i^c
+
\mathbf{T}_{\ell(i)}^{f}
-
\mathbf{g}_i^f
\right\|_2^2},
\end{equation}
where $\ell(i)$ denotes the cluster label of point $i$.

For completeness, given centered canonical and deformed neighborhoods, Kabsch forms
\begin{equation}
H_i = \sum_{j \in K_i}(x_c^j-\bar{x}_c)(x_f^j-\bar{x}_f)^T,
\end{equation}
computes $H_i=U\Sigma V^T$, and applies the determinant-corrected rotation $R_i=V\operatorname{diag}(1,1,\det(VU^T))U^T$.

\paragraph{Cluster Splitting.}
We iteratively split existing clusters with large reconstruction errors. For each split process, we select a seed point that simultaneously exhibits high reconstruction error and is far from the cluster centroid, encouraging separation of poorly explained and spatially distant regions. We initialize a new cluster from this seed point and its closest K (default K = 7) spatial neighboring points, while the remaining points stay in the original cluster. 

We then perform EM-style refinement: in the E-step, we estimate the rigid transformation of the temporal 
 point sequence in the new cluster; in the M-step, we compute the reconstruction error of all points under all clusters' transformation, and reassign each point to the cluster that best explains its motion according to the reconstruction error. The process, including split, compute transform, and point reassign, repeats until reaching the maximum number of clusters or until the overall reconstruction error no longer decreases after splitting, resulting in motion-consistent clusters with associated rigid transformations. See \cref{fig:motion cluster} as an example. Furthermore, a dynamic visualization can be seen in \suppl's live video.

\section{Curve Skeleton Design}
\label{sup:curve_skeleton}
Skinning weights implicitly encode joint locations~\cite{Example-based}, as joints can be interpreted as points that lie in the center of the boundary of the skinning weights. Concretely, given two spatially adjacent rigid parts with transformations $(R_1, T_1)$ and $(R_2, T_2)$ across frames $k$, the joint position $x$ can be estimated by minimizing the discrepancy between their transformed positions:
\[
\min_{x} \sum_{k} \left\| R_{1,k} x + T_{1,k} - \left( R_{2,k} x + T_{2,k} \right) \right\|^2.
\]

This formulation assumes that a joint should move consistently under both transformations, which works well for tube-like structures. However, for non-rigid objects (e.g., skirts), although SSDR can decompose surface points into rigid clusters and estimate joints, the resulting joint connectivity does not follow a tube-like kinematic structure but forms more complex structures (e.g., graph-like).

To address this, we introduce a curve skeleton as a geometric prior by mapping the object to an underlying tube-like structure, the curve skeleton. While the observed geometry may not follow a tube-like topology, we assume an intrinsic skeleton that is approximately tubular and rigid, such that joint estimation is restricted to lie on the curve skeleton rather than in the full 3D space.

\section{Inner Skeleton Acquisition}
\label{sup:inner_skeleton}

We denote the Gaussians in canonical space as $\mathcal{G}_c = \{\mathbf{g}_i^c\}$ and the curve skeleton as a graph $\mathcal{S}$ with nodes $\{s_j\}$ and connectivity edges.

Curve skeletonization establishes Gaussian-to-skeleton correspondence by mapping each $\mathbf{g}_i^c \in \mathcal{G}_c$ to its nearest skeleton point $s_{C_i}$ based on Euclidean distance in the canonical space. Then we transfer the skinning weights from Gaussians to the curve skeleton. To detect joints, we formulate a three-step process including weight aggregation, variation measurement, and joint selection:
\begin{equation}
\begin{aligned}
\mathcal{J} &= \left\{ s_j \in \mathcal{S} \mid \nabla W_j > \tau \right\}, \\
\nabla W_j &= \sum_{b=1}^{B} \left\| \nabla \bar{w}_{j,b} \right\|^2, 
\bar{w}_{j,b} = \frac{1}{|\mathcal{V}_j|} 
\sum_{\mathbf{g}_i^c \in \mathcal{V}_j} w_{i,b} 
\end{aligned}
\end{equation}
where $\mathcal{V}_j = \{ \mathbf{g}_i^c \mid C_i = j \}$ denotes the Gaussians associated with skeleton point $s_j$. To construct the kinematic tree, we use connectivity from the curve skeleton:
\begin{equation}
\begin{aligned}
\mathcal{E} = \bigl\{(s_j,s_k)\ \bigm|\ {} &(s_j,s_k)\text{ are connected in the curve skeleton},\\
&s_j,s_k\in\mathcal{J}\bigr\}.
\end{aligned}
\end{equation}

If the candidate graph contains a cycle, we remove cycle edges before hierarchy construction, selecting the loop-free structure that is most consistent with the observed part motion. A kinematic tree $\mathcal{T} = (\mathcal{J}, \mathcal{A})$ is then obtained by performing a depth-first search (DFS) traversal starting from the root joint, defined as the joint closest to the Gaussian center of mass in $\mathcal{G}_c$.

\section{Implementation Details}
\label{sup:implementation}
We implement $D_{\mathrm{MLP}}^{(\cdot)}(f)$ using an MLP with 8 linear layers and 256 feature dimensions. 
The reconstruction stage requires 40{,}000 iterations, taking approximately 15 minutes on a single RTX~4090D GPU. We optimize with Adam using a learning rate of $5\times10^{-6}$ and the default Adam settings. We set $\lambda_{\mathrm{ssim}}=0.5$, $\lambda_{\mathrm{ARAP}}=1$, and $\lambda_{\mathrm{dist}}=10$ for reconstruction.
The subsequent training-free stages (\cref{sec:skeletonization,sec:partmm})
, including motion clustering, SSDR, skeletonization, and PartMM, require only 2 minutes in total, demonstrating the computational efficiency of our pipeline. Detailed runtime analysis and profiling are provided in the \suppl (\cref{sup:runtime}).

For motion clustering, we set the maximum bone count to 50, use $K=7$ spatial neighbors per cluster for local-rigidity regularization, and set the skeletonization threshold of \cref{eq:joint_detection} to $\tau = 0.3$. In the matching stage, bones are partitioned into five overlapping parts, with each part performing $k$-nearest neighbor search ($k=7$). Matching operates across five pyramid levels to capture both coarse and fine motion patterns, with a default patch size of 7.

\section{DeformingThings4D Dataset Processing}
\label{sup:animal_processing}

The DeformingThings4D dataset comprises 57 animal categories, each containing multiple sequences with consistent mesh topology, totaling approximately 1000 frames per category. For each category, we first compute a motion magnitude score for every sequence, defined as the average Euclidean displacement of mesh vertices between consecutive frames, aggregated over the entire sequence. We then rank all sequences within the category based on this score and iteratively select full sequences in descending order of motion until the total number of selected frames exceeds 200. This strategy prioritizes highly dynamic sequences while preserving temporal continuity within each selected sequence.

We then perform SSDR optimization on the corresponding mesh sequences of the selected frames and extract the corresponding skeletons using our proposed skeleton extraction method. The resulting poses are further refined to improve temporal consistency and used as conditioning signals (\cref{sec:skeletonization}).

An 80/20 train-test split is applied at the frame level. MLP and GRU take the skeleton pose (joint rotations) as input and predict the 6-DOF transform parameters of outer bones. PartMM and FullMM construct a motion database from the training frames and retrieve or infer SSDR bones for test frames from this database, conditioned on the corresponding skeleton poses at test time.

\section{Dataset Configuration}
\label{sup:dataset_selection}

We adopt dataset-specific configurations for evaluation.

\noindent\textbf{DNA-Rendering.}
We use the standard 4K4D test setup.

\noindent\textbf{ActorsHQ.}
For each sequence, we compute a motion magnitude score based on the variation of the mesh bounding box across frames. We then select a 300-frame segment with the highest score and downsample it by a factor of 4 for evaluation.

Finally, all datasets use an 80/20 train-test split, and rendering metrics are computed on held-out novel poses.
\section{Runtime Analysis}
\label{sup:runtime}
\begin{figure}[h]
    \centering
    \includegraphics[width=\linewidth]{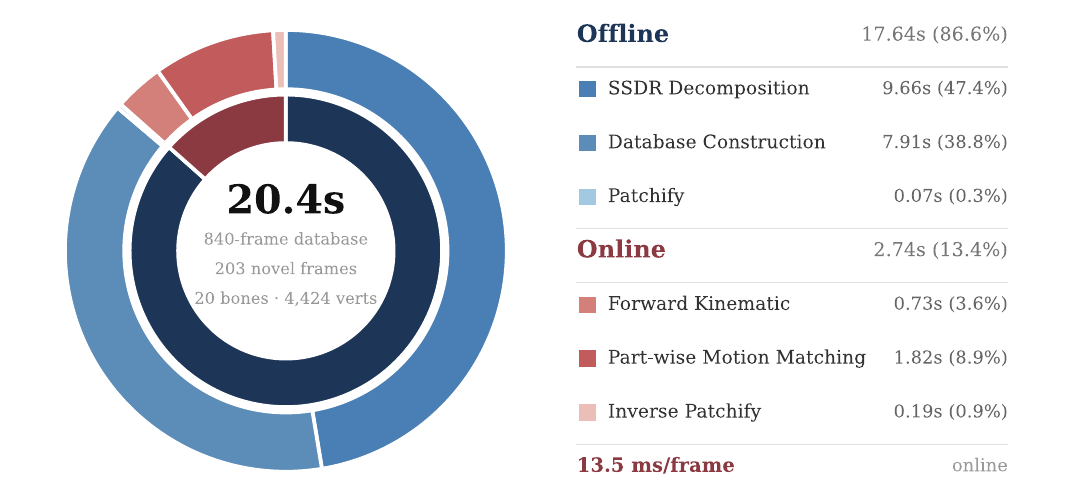}
    \caption{ \textbf{Runtime Analysis}.}
    \label{fig:runtime}
\end{figure}

We analyze runtime on VTO (T-shirt: 840 database + 203 query frames, 20 bones) with PartMM (\cref{fig:runtime}). Core runtime ($\sim$20 seconds) is much lower than $\sim$2 minutes in \cref{sec:exp_setting} since VTO provides SMPL skeletons. Overhead stems from data I/O and canonical skeleton computation ($\sim$1 minute); per-frame extraction dominates (>10 minutes). Temporal consistency constraints could mitigate this cost.
Online inference achieves 13.5\,ms per frame, enabling real-time control. Main bottlenecks: database construction and SSDR precomputation (addressable via learned motion matching), and forward kinematics (optimizable via CUDA batching).

\section{More Results}
\begin{figure}
    \centering
    \includegraphics[width=\linewidth]{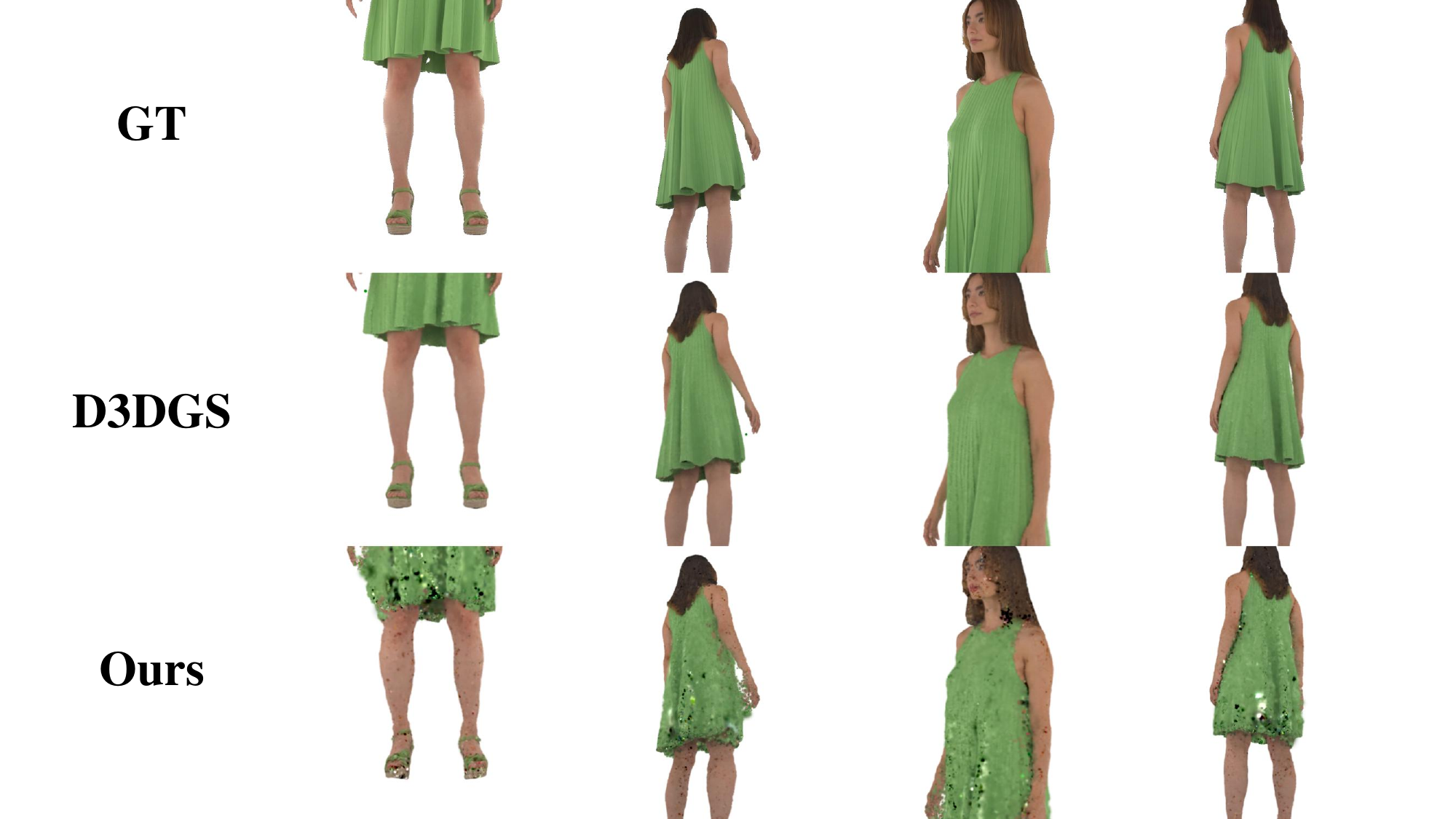}
    \caption{ \textbf{Rendering Artifacts Visualization} }
    \label{fig:render_issue}
\end{figure}

\begin{figure*}
    \centering
    \includegraphics[width=0.87\textwidth]{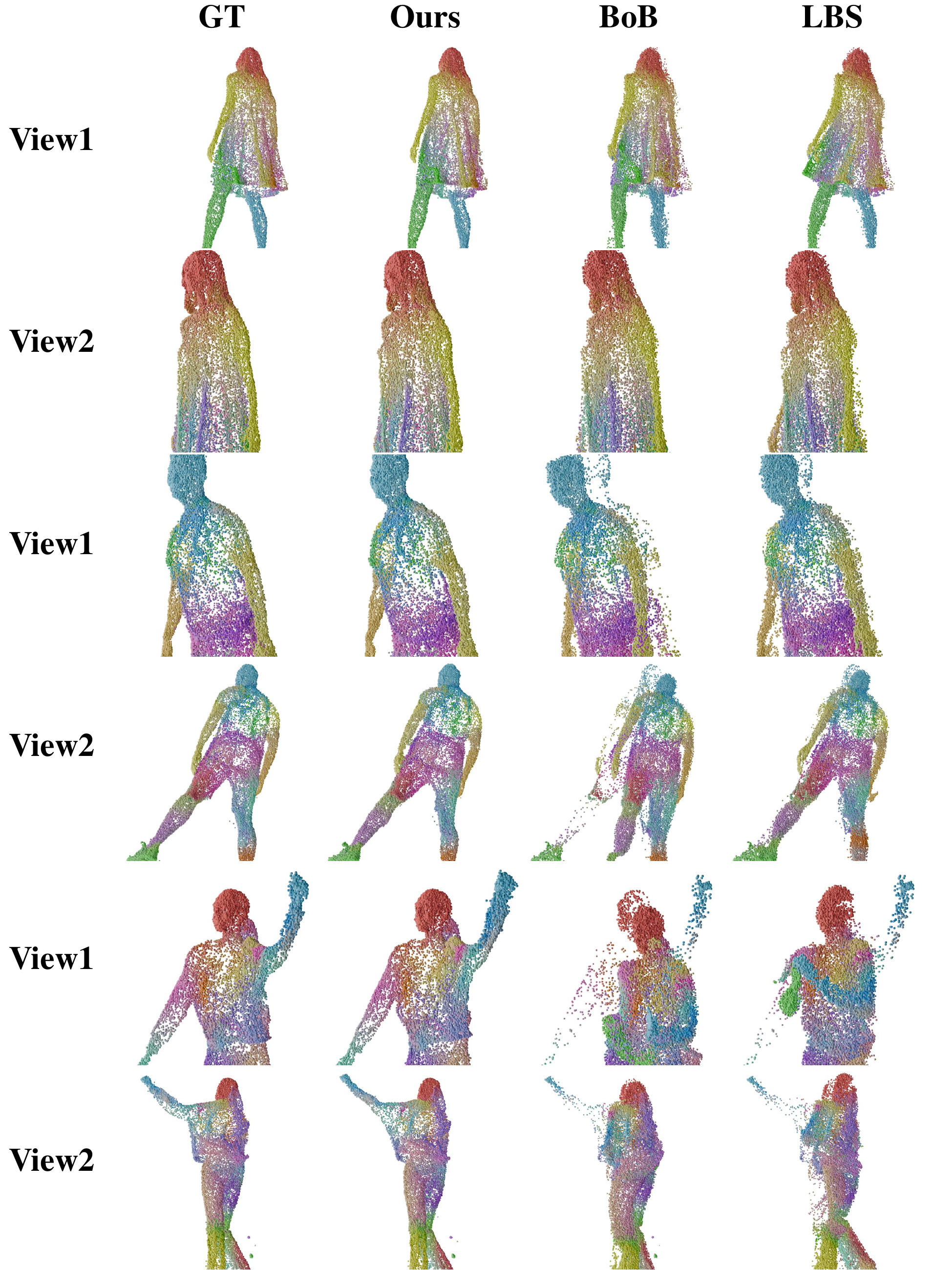}
    \caption{ \textbf{Rendering results of ActorsHQ Dataset} }
    \label{fig:actorshq}
\end{figure*}

\begin{figure*}[p]
    \centering
    \begin{minipage}[c][0.90\textheight][c]{\textwidth}
        \centering
        \begin{minipage}{\textwidth}
            \centering
            \includegraphics[width=0.82\linewidth,keepaspectratio]{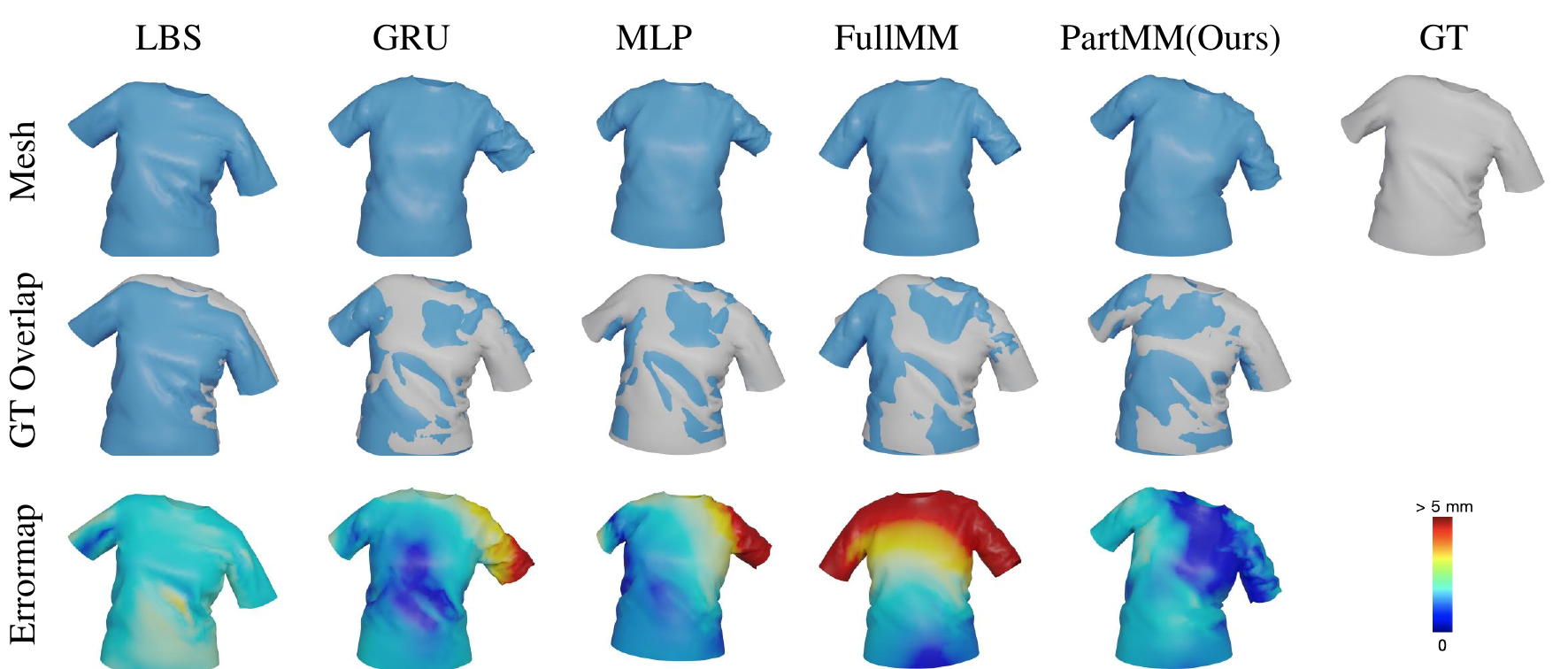}
            \captionof{figure}{\textbf{More Qualitative Comparison on VTO dataset}}
            \label{fig:vto_more2}
        \end{minipage}
        \par\vfill
        \begin{minipage}{\textwidth}
            \centering
            \includegraphics[width=0.82\linewidth,keepaspectratio,trim=0 7mm 0 0,clip]{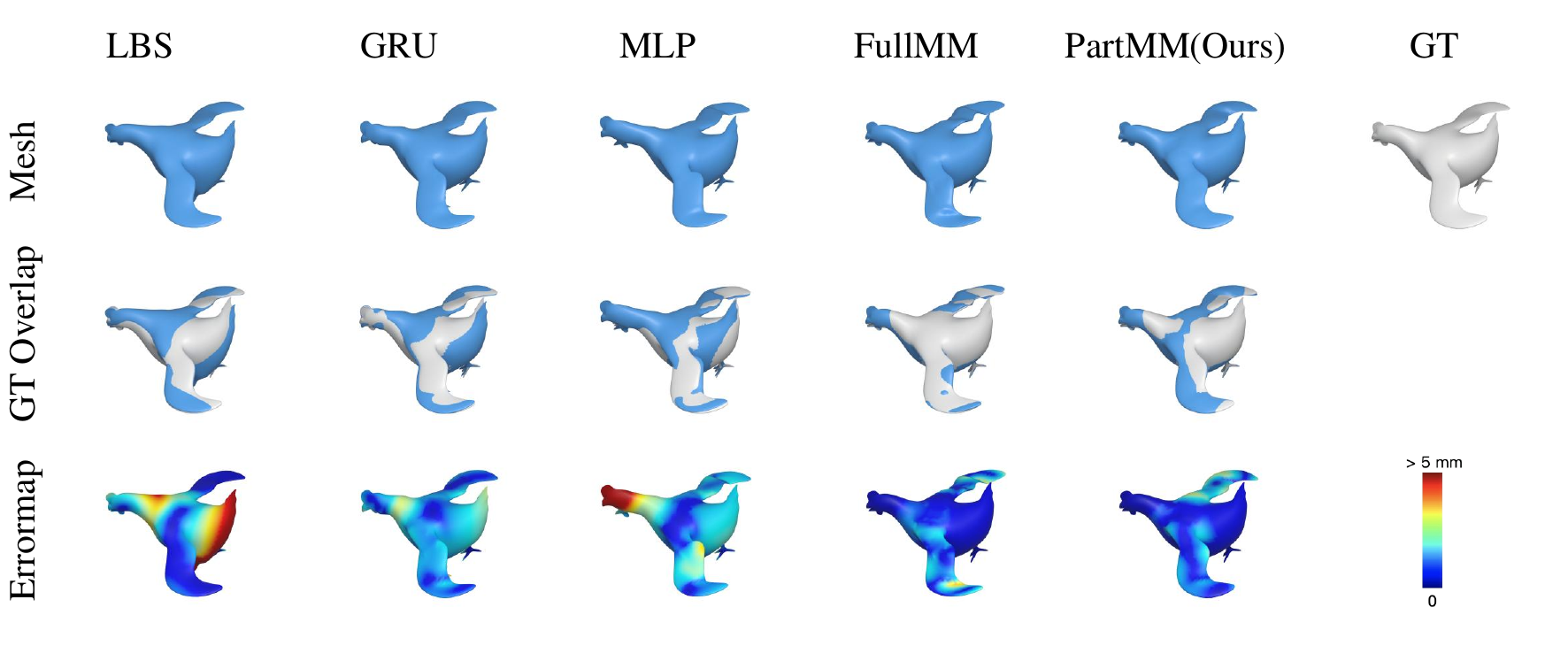}
            \captionof{figure}{\textbf{More Qualitative Comparison on D4D dataset}}
            \label{fig:animal_more}
        \end{minipage}
        \par\vfill
        \begin{minipage}{\textwidth}
            \centering
            \includegraphics[width=0.82\linewidth,keepaspectratio,trim=0 7mm 0 0,clip]{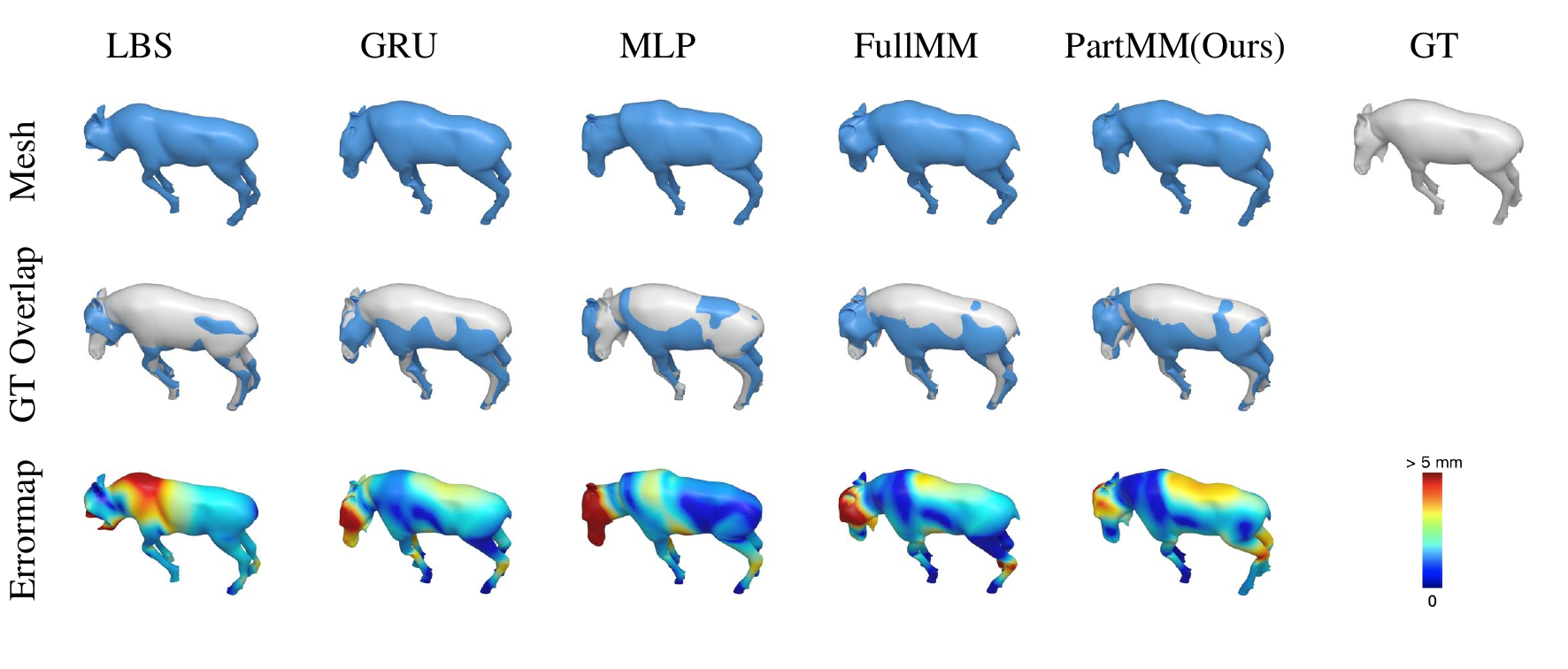}
            \captionof{figure}{\textbf{More Qualitative Comparison on D4D dataset}}
            \label{fig:animal_more2}
        \end{minipage}
    \end{minipage}
\end{figure*}

\begin{figure*}[p]
    \centering
    \begin{minipage}[c][0.85\textheight][c]{\textwidth}
        \centering
        \begin{minipage}{\textwidth}
            \centering
            \includegraphics[width=\linewidth,keepaspectratio,trim=0 7mm 0 0,clip]{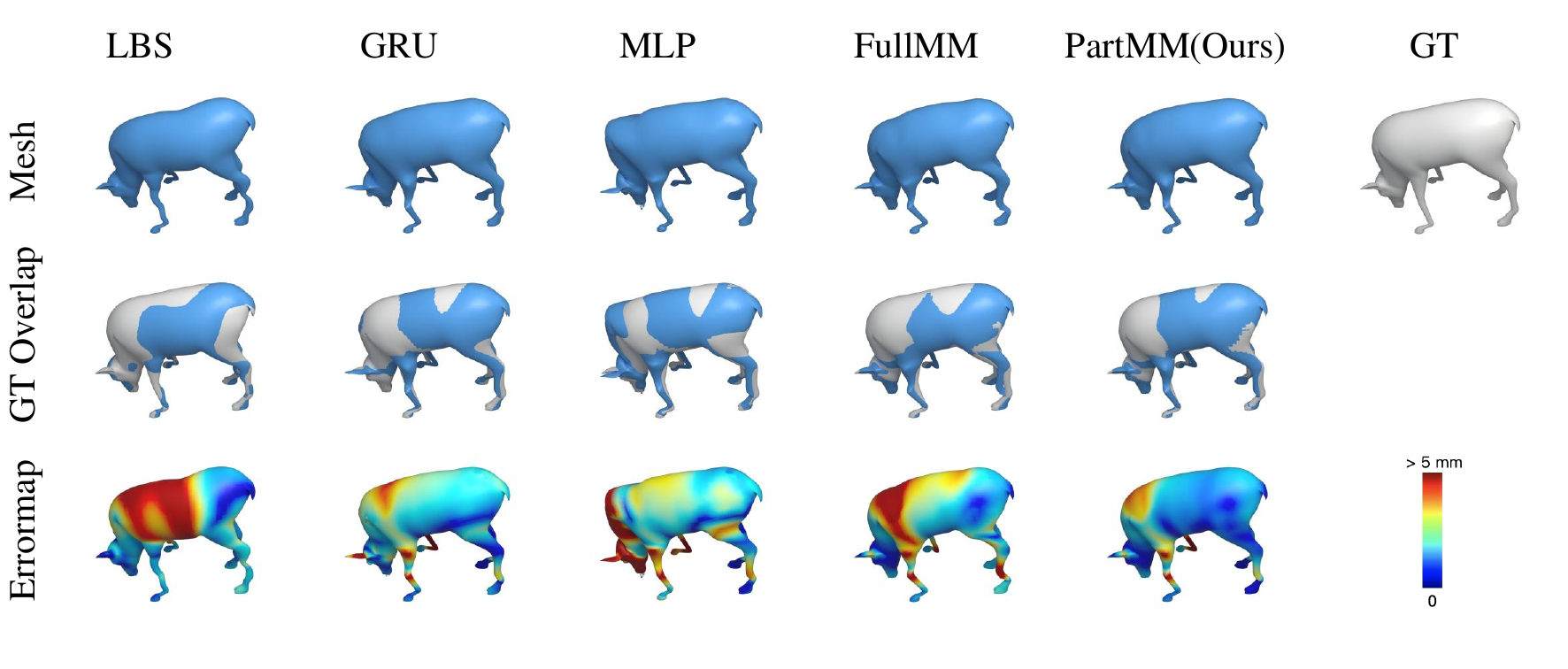}
            \captionof{figure}{\textbf{More Qualitative Comparison on D4D dataset}}
            \label{fig:animal_more3}
        \end{minipage}
        \par\vfill
        \begin{minipage}{\textwidth}
            \centering
            \includegraphics[width=\linewidth,keepaspectratio,trim=0 5mm 0 0,clip]{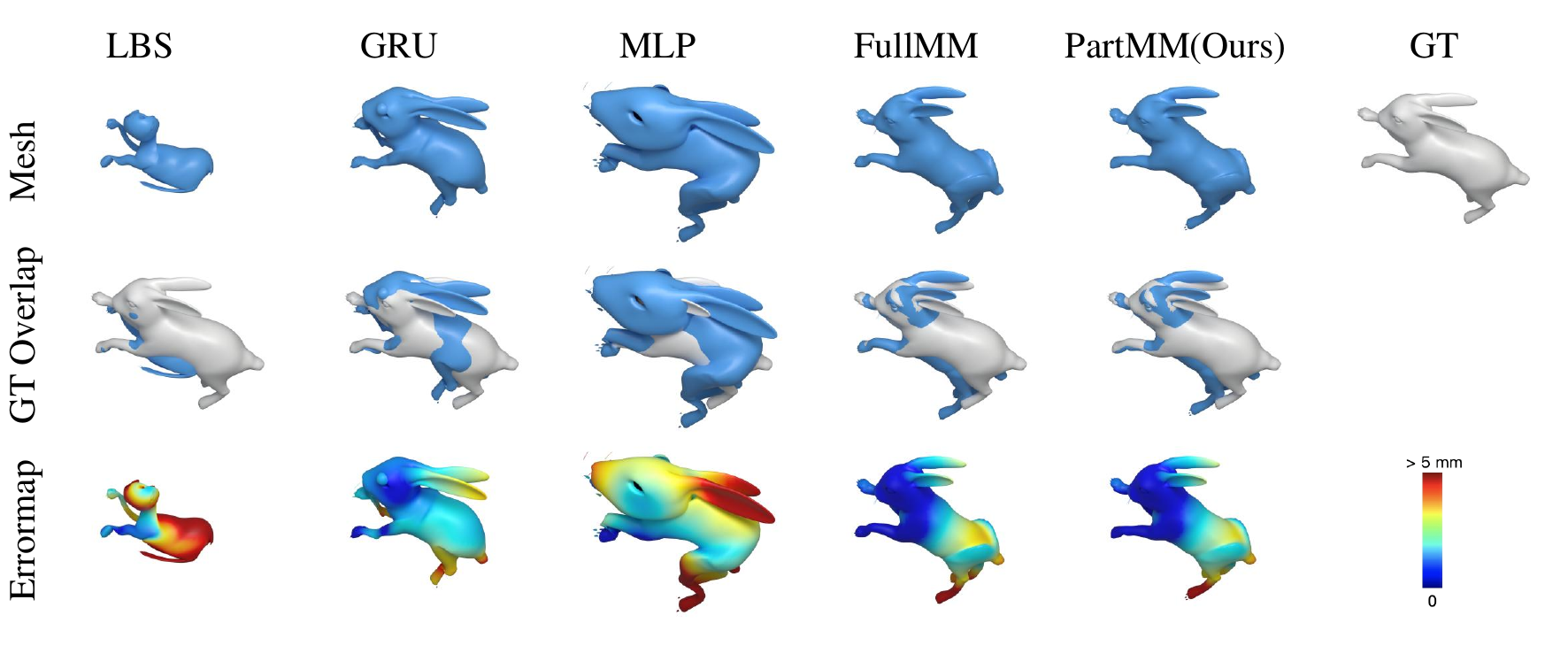}
            \captionof{figure}{\textbf{More Qualitative Comparison on D4D dataset}}
            \label{fig:animal_more4}
        \end{minipage}
    \end{minipage}
\end{figure*}

\Cref{fig:actorshq} and the video show ActorsHQ results with stronger geometric guidance and fewer artifacts. Geometry-only driving lacks appearance correction, so image-space refinement~\cite{li2024animatablegaussians} may help; imperfect masks can also cause black regions (\cref{fig:render_issue}). Synthetic results and the RigGS skeleton comparison are included in the video for completeness and qualitative inspection.

\end{subappendices}

\end{document}